\documentclass[5p]{elsarticle}
\usepackage{amsmath,amssymb,amsfonts}
\usepackage{paralist}                      
\usepackage{bm}
\usepackage{wrapfig}
\usepackage{float}
\usepackage{multirow}
\usepackage{booktabs}
\usepackage{nicefrac}
\usepackage{nth}
\usepackage[bf,small,skip=0pt]{caption}
\usepackage{subcaption}
\usepackage{wrapfig}
\usepackage{url}
\usepackage{dsfont}
\usepackage{nicefrac}
\usepackage[table]{xcolor}
\usepackage[colorlinks=true]{hyperref}
\usepackage[draft]{fixme}
\usepackage{tabularx,colortbl}
\usepackage{makecell}
\usepackage{fixme}

\journal{Neuroimage}

\newcommand{\xy}[1]{{\color{black}{#1}}}
\newcommand{\mn}[1]{{\color{black}{#1}}}

\newcommand{\mnr}[1]{{\color{black}{#1}}}
\newcommand{\xyadd}[1]{{\color{black}{#1}}}

\newcommand{\xyrevision}[1]{{\color{black}{#1}}}
\newcommand{\xyrevisiondl}[1]{{\color{black}{#1}}}
\newcommand{\xyrevisionsecond}[1]{{\color{black}{#1}}}

\newcommand{\rknew}[1]{{\color{black}{#1}}}

\newcommand{\mnminor}[1]{{\color{black}{#1}}}

\newcolumntype{Y}{>{\centering\arraybackslash}X}

\begin{document}

\begin{frontmatter}

\title{\texttt{Quicksilver}: Fast Predictive Image Registration \\ -- a Deep Learning Approach}

\author{Xiao Yang$^\dagger$, Roland Kwitt$^+$, Martin Styner$^{\dagger, \$}$ and Marc Niethammer$^{\dagger,*}$}
\address{$^\dagger$University of North Carolina at Chapel Hill, Chapel Hill, USA\\
  $^*$Biomedical Research Imaging Center (BRIC), Chapel Hill, USA\\
  $^\$$Department of Psychiatry, UNC Chapel Hill, USA\\
$^+$Department of Computer Science, University of Salzburg, Austria}

\begin{abstract}
  This paper introduces \texttt{Quicksilver}, a fast deformable image registration method.  \texttt{Quicksilver} registration for image-pairs works by patch-wise prediction of a deformation model based {\it directly} on image appearance. A deep encoder-decoder network is used as the prediction model. While the prediction strategy is general, we focus on predictions for the Large Deformation Diffeomorphic Metric Mapping (LDDMM) model. Specifically, we predict the momentum-parameterization of LDDMM, which facilitates a patch-wise prediction strategy while maintaining the theoretical properties of LDDMM, such as guaranteed diffeomorphic mappings for sufficiently strong regularization. We also provide a probabilistic version of our prediction network which can be sampled during the testing time to calculate uncertainties in the predicted deformations. Finally, we introduce a new correction network which greatly increases the prediction accuracy of an already existing prediction network. \mnr{We show experimental results for uni-modal atlas-to-image as well as uni- / multi-modal} image-to-image registrations. These experiments \rknew{demonstrate} that our method accurately predicts registrations obtained by numerical optimization, is very fast, achieves state-of-the-art registration results on four standard validation datasets\mnr{, and can jointly learn an image similarity measure.} \texttt{Quicksilver} is freely available as an open-source software.
\end{abstract}


\end{frontmatter}

\section{Introduction}
Image registration is a key component for medical image analysis to provide spatial correspondences. Image registration is typically formulated as an optimization problem~\cite{modersitzki2004}, optimizing the parameters of a transformation model. The goal is to achieve the best possible agreement between a transformed source and a target image, subject to transformation constraints. Apart from simple\rknew{,} low-dimensional parametric models (e.g., rigid or affine transformations), more complex, high-dimensional parametric or non-parametric registration models are able to capture subtle, localized image deformations. However, these methods, in particular, the non-parametric approaches, have \rknew{a} very large numbers of parameters. Therefore, numerical optimization to solve the registration problems becomes computationally costly, even with acceleration by graphics processing units (GPUs).

\vskip1ex
While computation time may not be overly critical for imaging studies of moderate size, rapid registration approaches are needed to (i) allow for interactive analysis, to (ii) allow their use as building blocks for more advanced image analysis algorithms; and to (iii) time- and cost-efficiently analyze very large imaging studies. As a case in point, sample sizes of neuroimaging studies are rapidly increasing. While, only two decades ago, neuroimaging studies with few tens of subjects were not unusual, we are now witnessing the emergence of truly large-scale imaging studies. For example, the UK Biobank study is, at the moment, the world's largest health imaging study and will image ``the brain, bones, heart, carotid arteries and abdominal fat of 100,000 participants'' using magnetic resonance (MR) imaging within the next few years~\cite{biobankWebsite}. {\it Furthermore, image sizes are increasing drastically}. While, a decade ago, structural MR images of human brains with voxel sizes of $2\times 2 \times 2~$mm$^3$ were typical for state-of-the-art MR acquisitions, today we have voxel sizes smaller than $1\times 1\times 1~$mm$^3$ as, for example, acquired by the human connectome project~\cite{vanessen2013}. This increase in image resolution increases the data size by an order of magnitude. Even more dramatically, the microscopy field now routinely generates gigabytes of high-resolution imaging data, for example, by 3D imaging via tissue clearing~\cite{chung2013clarity}. Hence, fast, memory-efficient, and parallelizable image analysis approaches are critically needed. In particular, such approaches are needed for deformable image registration, which is a key component of many medical image analysis systems. 

\vskip1ex
Attempts at speeding-up deformable image registration have primarily focused on GPU implementations~\cite{shams2010survey}, with impressive speed-ups over their CPU-based counterparts. However, these approaches are still relatively slow. Runtimes in the tens of minutes are the norm for popular deformable image registration \rknew{solutions}. For example, a GPU-based registration of a \xyrevision{$128\times128\times128$} image volume using LDDMM will take about 10 minutes on a current GPU (e.g., a Nvidia TitanX). This is much too slow to allow for large-scale processing, the processing of large datasets, or close to interactive registration tasks. Hence, improved algorithmic approaches are desirable. Recent work has focused on \mnr{{\it better numerical methods} and} {\it approximate} approaches. For example, \mnr{Ashburner and Friston~\cite{ashburner2011} use a Gauss--Newton method to accelerate convergence for LDDMM and} Zhang et al.~\cite{zhang2015} propose a finite-dimensional approximation of LDDMM, achieving a roughly $25\times$ speed-up over a standard LDDMM optimization-based solution.

\vskip1ex
An alternative approach to improve registration speed is to {\it predict} deformation parameters, or deformation parameter update steps in the optimization via a regression model, instead of directly minimizing a registration energy~\cite{gutierrez2017,gutierrez2016learning,chou20132d}. The resulting predicted deformation fields can either be used directly, or as an initialization of a subsequent optimization-based registration. However, the high dimensionality of the deformation parameters as well as the non-linear relationship between the images and the parameters pose a significant challenge. Among these methods, Chou et al.~\cite{chou20132d} propose a multi-scale linear regressor which only applies to affine deformations and low-rank approximations of non-linear deformations. Wang et al.~\cite{Wang201561} predict deformations by key-point matching using sparse learning followed by dense deformation field generation with radial basis function interpolation. The performance of the method heavily depends on the accuracy of the key point selection. Cao et al.~\cite{Tian2015} use a semi-coupled dictionary learning method to directly model the relationship between the image appearance and the deformation parameters of the LDDMM model~\cite{beg2005}. However, only a linear relationship is assumed between image appearance and the deformation parameters. Lastly, Gutierrez et al.~\cite{gutierrez2016learning} use a regression forest \mnr{and gradient boosted trees~\cite{gutierrez2017}} based on hand-crafted features to learn update steps for a \mnr{rigid and a} B-spline registration model. 

\vskip1ex
In this work, we propose a deep regression model to predict deformation parameters using image appearances in a time-efficient manner. Deep learning has been used for optical flow estimation~\cite{deepflow, flownet} and deformation parameter prediction for affine transformations~\cite{Miao2016}. We investigate a non-parametric image registration approach, where we predict voxel-wise deformation parameters from image patches. Specifically, we focus on the initial momentum LDDMM shooting model~\cite{Vialard2012}, as it has many desirable properties: 
\begin{itemize}
  \item It is based on Riemannian geometry, and hence induces a distance metric on the space of images. 
  \item It can capture large deformations. 
  \item It results in highly desirable diffeomorphic spatial transformations (if regularized sufficiently). I.e., transformations which are smooth, one-to-one and have a smooth inverse. 
  \item It uses the initial momentum as the registration parameter, which does not need to be spatially smooth, and hence can be predicted patch-by-patch, and from which the whole geodesic path can be computed. 
\end{itemize}
The LDDMM shooting model \mnr{in} of itself is important for various image analysis tasks such as principal component analysis~\cite{Vaillant2004S161} and image regression~\cite{Niethammer2011, Yi2012}.


\vskip1ex
\noindent
Our \emph{contributions} are as follows:
\begin{itemize}
  \item {\it Convenient parameterization:} Diffeomorphic transformations are desirable in medical image analysis applications to smoothly map between fixed and moving images, or to and from an atlas image. Methods, such as LDDMM, with strong theoretical guarantees exist, but are typically computationally very demanding. On the other hand, direct prediction, e.g., of optical flow~\cite{deepflow,flownet}, is fast, but the regularity of the obtained solution is unclear as it is not considered within the regression formulation. We demonstrate that the momentum-parameterization for LDDMM shooting~\cite{Vialard2012} is a convenient representation for regression approaches as (i) the momentum is typically compactly supported around image edges and (ii) there are no smoothness requirements on the momentum itself. Instead, smooth velocity fields are obtained in LDDMM from the momentum representation by {\it subsequent} smoothing. Hence, by predicting the momentum, we retain all the convenient mathematical properties of LDDMM and, at the same time, are able to predict diffeomorphic transformations {\it fast}. As the momentum has compact support around image edges, no ambiguities arise within uniform image areas (in which predicting a velocity or deformation field would be difficult). 
  \item {\it Fast computation:} We use a sliding window to locally predict the LDDMM momentum from image patches. We experimentally show that by using patch pruning and a large sliding window stride, our method achieves dramatic speedups compared to the optimization approach, while maintaining good registration accuracy.
  \item {\it Uncertainty quantification:} We extend our network to a Bayesian model which is able to determine the uncertainty of the registration parameters and, as a result, the uncertainty of the deformation field. This uncertainty information could be used, e.g., for uncertainty-based smoothing~\cite{simpson2011longitudinal}, or for surgical treatment planning, or could be directly visualized for qualitative analyses.
  \item {\it Correction network:} Furthermore, we propose a correction network to increase the accuracy of the prediction network. Given a trained prediction network, the correction network predicts the difference between the ground truth momentum and the predicted result. The difference is used as a correction to the predicted momentum to increase prediction accuracy. Experiments show that the correction network improves registration results to the point where optimization-based and predicted registrations achieve a similar level of registration accuracy on registration validation experiments.
  \item \xyrevision{{\it Multi-modal registration:} We also explore the use of our framework for multi-modal image registration prediction. The goal of multi-modal image registration is to establish spatial correspondences between images acquired by different modalities. Multi-modal image registration is, \rknew{in general}, significantly more difficult than uni-modal image registration since image appearance can change drastically between different modalities. General approaches address multi-modal image registration by either performing image synthesis~\cite{cao2014,wein2008} to change the problem to an uni-modal image registration task, or by proposing complex, hand-crafted~\cite{viola1997, Meyer1998, Hermosillo2002, Lorenzen2006} or learned~\cite{Guetter2005,Lee2009,Michel2011,Cheng2015, simonovsky2016} multi-modal image similarity measures. In contrast, we \rknew{demonstrate} that our framework can {\it simultaneously} predict registrations and learn a multi-modal image similarity measure. Our experiments show that our approach also predicts accurate deformations for multi-modal registration. }
  \item {\it Extensive validation:} We extensively validate our predictive image registration approach for uni-modal image registration on the four validation datasets of Klein et al.~\cite{Klein2009786} and demonstrate registration accuracies on these datasets on par with the state-of-the-art. \mnr{Of note, these registration results are achieved using a model that was trained on an entirely different dataset (images from the OASIS dataset). Furthermore, we validate our model trained for multi-modal image registration using the IBIS 3D dataset~\cite{hazlett2017early}. Overall, our results are based on more than 2,400 image registration pairs.}
\end{itemize}
The registration method described here, which we name \texttt{Quicksilver}, is an extension of the preliminary ideas we presented in a recent workshop paper~\cite{YangFast2016} \mnr{and in a conference paper~\cite{yang2017multimodal}}. This paper offers more details of our proposed approaches, introduces the idea of improving registration accuracy via a correction network, and includes a comprehensive set of experiments for image-to-image registration.

\vskip1ex
\noindent
\textbf{Organization.}
The remainder of the paper is organized as follows. Sec.~\ref{sec:lddmm} reviews the registration parameterization of the shooting-based LDDMM registration algorithm. Sec.~\ref{sec:predict} introduces our deep network architecture for deformation parameter prediction, the Bayesian formulation of our network, as well as our strategy for speeding up the deformation prediction. Sec.~\ref{sec:correct} discusses the {\it correction network} and the reason why it improves the registration prediction accuracy over an existing prediction network. Sec.~\ref{sec:experiment} presents experimental results for \emph{atlas-to-image} and \emph{image-to-image} registration. Finally, Sec.~\ref{sec:discussion} discusses potential extensions and applications of our method.
\section{\xyrevisionsecond{Materials and Methods}}
\subsection{LDDMM Shooting}
\label{sec:lddmm}
Given a moving (source) image $M$ and a target image $T$, the goal of image registration is to find a deformation map $\Phi:\mathbb{R}^d \rightarrow \mathbb{R}^d$, which maps the moving image to the target image in such a way that the deformed moving image is similar to the target image, i.e., $M\circ\Phi^{-1}(x)\approx T(x)$. Here, $d$ denotes the spatial dimension and $x$ is the spatial coordinate of the fixed target image $T$. \mnr{Due to the importance of image registration, a large number of different approaches have been proposed~\cite{modersitzki2004,hill2001,sotiras2013,oliveira2014}. Typically, these approaches are formulated as optimization problems, where one seeks to minimize an energy of the form
  \begin{equation}
    E(\Phi) = {\rm Reg}[\Phi] + \frac{1}{\sigma^2}{\rm Sim}[I_0\circ\Phi^{-1},I_1],
  \end{equation}
  where $\sigma>0$ is a balancing constant, ${\rm Reg}[\cdot]$ regularizes the spatial transformation, $\Phi$, by penalizing spatially irregular (for example non-smooth) spatial transformations, and ${\rm Sim}[\cdot,\cdot]$ is an image dissimilarity measure, which becomes small if images are similar to each other. Image dissimilarity is commonly measured by computing the sum of squared differences (SSD) between the warped source image ($I_0\circ\Phi^{-1}$) and the target image ($I_1$), or via (normalized) cross-correlation, or mutual information~\cite{Hermosillo2002,modersitzki2004}. For simplicity, we use SSD in what follows, but other similarity measures could also be used. The regularizer ${\rm Reg}[\cdot]$ encodes what should be considered a plausible spatial transformation\footnote{\mnr{A regularizer is not necessarily required for simple, low-dimensional transformation models, such as rigid or affine transformations.}}. The form of the regularizer depends on how a transformation is represented. In general, one distinguishes between parametric and non-parametric transformation models~\cite{modersitzki2004}. Parametric transformation models make use of a relatively low-dimensional parameterization of the transformation. Examples are rigid, similarity, and affine transformations. But also the highly popular B-spline models~\cite{rueckert1999} are examples of parametric transformation models. Non-parametric approaches on the other hand parameterize a transformation locally, with a parameter (or parameter vector) for each voxel. The most direct non-parametric approach is to represent voxel displacements, $u(x) = \Phi(x)-x$. Regularization then amounts to penalizing norms involving the spatial derivatives of the displacement vectors. Regularization is necessary for non-parametric approaches to avoid ill-posedness of the optimization problem. Optical flow approaches, such as the classical Horn and Schunck optical flow~\cite{horn1981}, the more recent total variation approaches~\cite{zach2007}, or methods based on linear elasticity theory~\cite{modersitzki2004} are examples for displacement-based registration formulations. Displacement-based approaches typically penalize large displacements strongly and hence have difficulty capturing large image deformations. Furthermore, they typically also only offer limited control over spatial regularity. Both shortcomings can be circumvented. The first by applying greedy optimization strategies (for example, by repeating registration and image warping steps) and the second, for example, by explicitly enforcing image regularity by constraining the determinant of the Jacobian of the transformation~\cite{haber2007}. An alternative approach to allow for large deformations, while assuring diffeomorphic transformations, is to parameterize transformations via static or time-dependent velocity fields~\cite{vercauteren2009,beg2005}. In these approaches, the transformation $\Phi$ is obtained via time integration. For sufficiently regular velocity fields, diffeomorphic transformations can be obtained. As the regularizer operates on the velocity field(s) rather than the displacement field, large deformations are no longer strongly penalized and hence can be captured.
}

\mnr{LDDMM is a non-parametric registration method which represents the transformation via spatio-temporal velocity fields. In particular,} the sought-for mapping, $\Phi$, is obtained via an integration of a spatio-temporal velocity field $v(x,t)$ for unit time, \xyrevision{where $t$ indicates time and $t\in[0, 1]$}, such that $\Phi_t(x,t) = v(\Phi(x,t),t)$ and the sought-for mapping is $\Phi(x,1)$. To single-out desirable velocity-fields, non-spatial-smoothness at any given time $t$ is penalized \mnr{by the regularizer ${\rm Reg}[\cdot]$, which is applied to the velocity field instead of the transform $\Phi$ directly}. Specifically, LDDMM aims at minimizing the energy\footnote{When clear from the context, we suppress spatial dependencies for clarity of notation and only specify the time variable. E.g., we write $\Phi^{-1}(1)$ to mean $\Phi^{-1}(x,1)$.}~\cite{beg2005}
\begin{multline}
  E(v) = \int_0^1 \|v\|_L^2~dt + \frac{1}{\sigma^2}\|M\circ\Phi^{-1}(1)-T\|^2,\\ \text{s.t.}\quad \Phi_t(x,t) = v(\Phi(x,t),t),~\Phi(x,0)=\text{id}\label{eq:relaxation}
\end{multline}
where $\sigma>0$, $\|v\|_L^2 = \langle L v, v\rangle$, $L$ is a self-adjoint differential operator\footnote{Note that we define $\|v\|_L^2$ here as $\langle L v,v\rangle$ instead of $\langle Lv,Lv\rangle = \langle L^\dagger L v, v\rangle$ as for example in Beg et al.~\cite{beg2005}.}, $\text{id}$ is the identity map, and the differential equation constraint for $\Phi$ can be written in Eulerian coordinates as $\Phi^{-1}_t + D\Phi^{-1}v=0$, where \xyrevision{$\Phi_t(x,t)$ is the derivative of $\Phi$ with respect to time $t$, and} $D$ is the Jacobian matrix. In this LDDMM formulation (termed the {\it relaxation} formulation as a geodesic path -- the optimal solution -- is only obtained at optimality) the registration is parameterized by the full spatio-temporal velocity field $v(x,t)$. \mnr{From the perspective of an individual particle, the transformation is simply obtained by following the velocity field over time. To optimize over the spatio-temporal velocity field one solves the associated adjoint system backward in time, where the final conditions of the adjoint system are determined by the current image mismatch as measured by the chosen similarity measure~\cite{beg2005}. This adjoint system can easily be determined via a constrained optimization approach~\cite{hart2009} (see~\cite{borzi2003} for the case of optical flow). From the solution of the adjoint system one can compute the gradient of the LDDMM energy with respect to the velocity field at any point in time\footnote{\mnr{This approach is directly related to what is termed error backpropagation in the neural networks community~\cite{lecun88} as well as the reverse mode in automatic differentiation~\cite{griewank2008}. The layers in neural networks are analogous to discretized time-steps for LDDMM. The weights which parameterize a neural network are analogous to the velocity fields for LDDMM. Error-backpropagation via the chain rule in neural networks corresponds to the adjoint system in LDDMM, which is a partial differential equation when written in the Eulerian form in the continuum.}} and use it to numerically solve the optimization problem, for example, by a line-search~\cite{nocedal2006}. At convergence, the optimal solution will fulfill the optimality conditions of the constrained LDDMM energy of Eq.~\eqref{eq:relaxation}. These optimality conditions can be interpreted as the continuous equivalent of the Karush-Kuhn-Tucker conditions of constrained optimization~\cite{nocedal2006}. On an intuitive level, if one were to find the shortest path between two points, one would (in Euclidean space) obtain the straight line connecting these two points. This straight line is the geodesic path in Euclidean space. For LDDMM, one instead tries to find the shortest path between two images based on the minimizer of the inexact matching problem of Eq.~\eqref{eq:relaxation}. The optimization via the adjoint equations corresponds to starting with a possible path and then successively improving it, until the optimal path is found. Again, going back to the example of matching points, one would start with any possible path connecting the two points and then successively improve it. The result at convergence is the optimal straight line path.
  }

\mnr{Convergence to the shortest path immediately suggests an alternative optimization formulation. To continue the point matching example: if one knows that the optimal solution needs to be a straight line (i.e., a geodesic) one can consider optimizing only over the space of straight lines instead of all possible paths connecting the two points. This dramatically reduces the parameter space for optimization as one now only needs to optimize over the y-intercept and the slope of the straight line.} \mnr{LDDMM can also be formulated in such a way. One obtains the {\it shooting} formulation~\cite{Vialard2012,Niethammer2011}, which parameterizes the deformation via the initial momentum vector field $m_0=m(0)$ and the initial map $\Phi^{-1}(0)$, from which the map $\Phi$ can be computed for any point in time. The initial momentum corresponds to the slope of the line and the initial map corresponds to the y-intercept. The geodesic equations correspond to the line equation. The geodesic equations, in turn, correspond to the optimality conditions of Eq.~\eqref{eq:relaxation}.} Essentially, the shooting formulation enforces \mnr{these} optimality conditions of Eq.~\eqref{eq:relaxation} as a constraint. In effect, one then searches only over geodesic paths, as these optimality conditions are geodesic equations. They can be written in terms of the momentum $m$ alone. In particular, the momentum is the dual of the velocity $v$, which is an element in the reproducing kernel Hilbert space $V$; $m$ and $v$ are connected by a positive-definite, self-adjoint differential smoothing operator $K$ by $v = Km$ and $m = Lv$, where $L$ is the inverse of $K$. Given $m_0$, the complete spatio-temporal deformation $\Phi(x,t)$ is determined.

Specifically, the energy to be minimized for the shooting formulation of LDDMM is~\cite{Singh2013}
\begin{equation}
    E(m_0) = \langle m_0, Km_0\rangle + \frac{1}{\sigma^2}||M\circ \Phi^{-1}(1) - T||^2,\quad \text{s.t.}
    \label{eqn:momentum_energy}
\end{equation}
\begin{equation}
\begin{split}
m_t + \text{ad}_{v}^*m = 0,\\ m(0)=m_0, \\ \Phi^{-1}_t + D \Phi^{-1} v = 0,\\  \Phi^{-1}(0)=\text{id}, \\ m - Lv = 0\enspace,
\label{eqn:forward}
\end{split}
\end{equation}
\noindent where id is the identity map, and the operator $ad^{*}$ is the dual of the negative Jacobi-Lie bracket of vector fields, i.e., $\text{ad}_v w = -[v,w] = Dvw - Dwv$. \mnr{The optimization approach is similar to the one for the relaxation formulation. I.e., one determines the adjoint equations for the shooting formulation and uses them to compute the gradient with respect to the unknown initial momentum $m_0$~\cite{Singh2013,Vialard2012}. Based on this gradient an optimal solution can, for example, be found via a line-search or by a simple gradient descent scheme.}

\begin{figure*}
  \centering
  \begin{tabular}{>{\centering\arraybackslash}m{0.72\textwidth}|>{\centering\arraybackslash}m{0.24\textwidth}} 
    \includegraphics[width=0.7\textwidth]{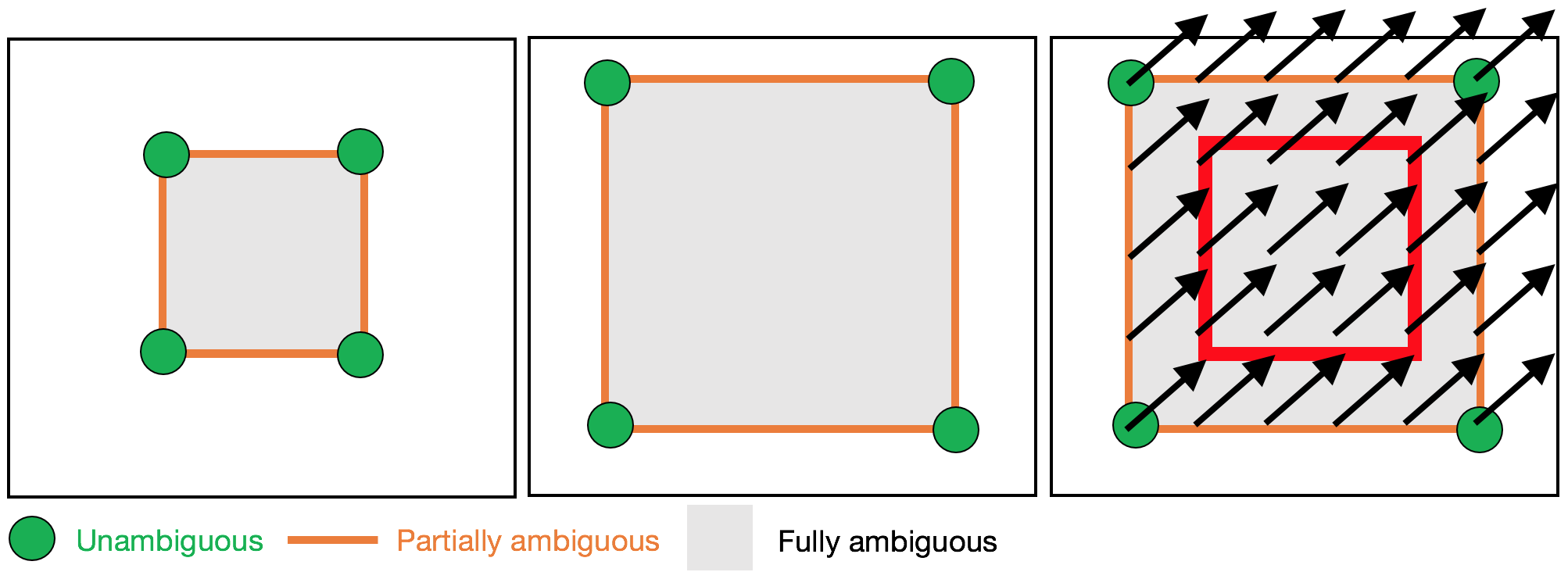} &
    \includegraphics[width=0.22\textwidth]{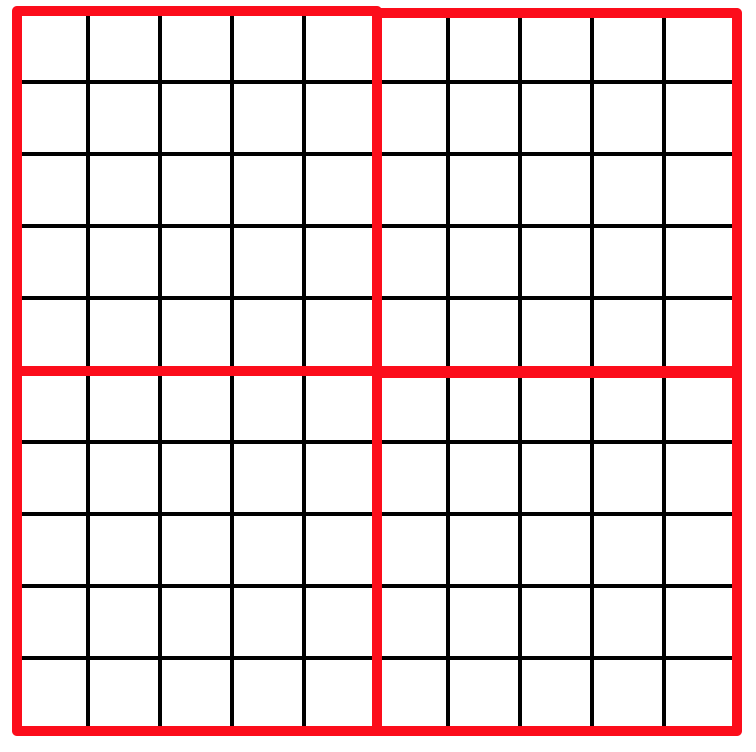}
  \end{tabular}
  \caption{{\bf Left:} The LDDMM momentum parameterization is ideal for patch-based prediction of image registrations. Consider registering a small \mnr{square (left)} to a large square \mnr{(middle)} with uniform intensity. Only the corner points suggest clear spatial correspondences. Edges also suggest spatial correspondences, however, correspondences between {\it individual} points on edges remain ambiguous. Lastly, points interior to the squares have ambiguous spatial correspondences, which are established purely based on regularization. Hence, predicting velocity or displacement fields (which are spatially dense) from patches is challenging in these interior areas \mnr{(right)}, in the absence of sufficient spatial context. \mnr{Predicting a displacement field as illustrated in the right image from an interior patch (illustrated by the red square) would be impossible if both the target and the source image patches are uniform in intensity. In this scenario, the patch information would not provide sufficient spatial context to capture aspects of the deformation.} On the other hand, we know from LDDMM theory that the optimal momentum, $m$, to match images can be written as $m(x,t)=\lambda(x,t)\nabla I(x,t)$, where $\lambda(x,t)\mapsto\mathbb{R}$ is a spatio-temporal scalar field and $I(x,t)$ is the image at time $t$~\cite{hart2009,Niethammer2011,Vialard2012}. Hence, in spatially uniform areas (where correspondences are ambiguous) $\nabla I=0$ and consequentially $m(x,t)=0$. This is highly beneficial for prediction as the momentum only needs to be predicted at image edges. {\bf Right:}  Furthermore, as the momentum is not spatially smooth, the regression approach does not need to account for spatial smoothness, which allows predictions with non-overlapping or hardly-overlapping patches \mnr{as illustrated in the figure by the red squares}. This is not easily possible for the prediction of displacement or velocity fields since these are expected to be spatially dense and smooth, which would need to be considered in the prediction. \mnr{Consequentially, predictions of velocity or displacement fields will inevitably result in discontinuities across patch boundaries (i.e., across the red square boundaries shown in the figure) if they are predicted independently of each other.}}
  \label{fig:registration_101}
\end{figure*}

\mnminor{A natural approach for deformation prediction would be to use the entire 3D moving and target images as input, and to directly predict the 3D displacement field. However, this is not feasible in our formulation (for large images) because of the limited memory in modern GPUs. We circumvent this problem by extracting image patches from the moving image and target image at the same location, and by then predicting deformation {\it parameters} for the patch. The entire 3D image prediction is then accomplished patch-by-patch via a sliding window approach. Specifically,} in our framework, we predict the initial momentum $m_0$ given the moving and target images patch-by-patch. Using the initial momentum for patch-based prediction is a convenient parameterization because (i) the initial momentum is generally not smooth, but is compactly supported at image edges and (ii) the initial velocity is generated by applying a smoothing kernel $K$ to the initial momentum. Therefore, the smoothness of the deformation does not need to be specifically considered during the parameter prediction step, but is imposed {\it after} the prediction. Since $K$ governs the theoretical properties or LDDMM, a strong $K$ assures diffeomorphic transformations\footnote{See~\cite{beg2005,dupuis1998} for the required regularity conditions.}, making predicting the initial momentum an ideal choice. However, predicting alternative parameterizations such as the initial velocity or directly the displacement field would make it difficult to obtain diffeomorphic transformations. Furthermore, it is hard to predict initial velocity or displacement for homogeneous image regions, as these regions locally provide no information from which to predict the spatial transformation. In these regions the deformations are purely driven by regularization. This is not a problem for the initial momentum parameterization, since the initial momentum in these areas, for image-based LDDMM, is zero. This can be seen as for image-based LDDMM~\cite{Vialard2012,Niethammer2011,hart2009} the momentum can be written as $m(x,t) = \lambda(x,t) \nabla I(x,t)$, where $\lambda$ is a scalar field and $\nabla I$ is the spatial gradient of the image. Hence, for homogeneous areas, $\nabla I =0$ and consequentially $m=0$. Fig.~\ref{fig:registration_101} illustrates this graphically. In summary, the initial momentum parameterization is ideal for our patch-based prediction method. \mnr{Note that since the initial momentum can be written as $m=\lambda\nabla I$ one can alternatively optimize LDDMM over the scalar-valued momentum $\lambda$. This is the approach that has historically been taken for LDDMM~\cite{beg2005,hart2009,Vialard2012}. However, optimizing over the vector-valued momentum, $m$, instead is numerically better behaved~\cite{Singh2013}, which is why we focus on it for our predictions. While we are not exploring the prediction of the scalar-valued momentum $\lambda$ here, it would be interesting to see how scalar-valued and vector-valued momentum predictions compare. In particular, since the prediction of the scalar-valued momentum would allow for simpler prediction approaches (see details in Sec.~\ref{sec:predict}).}
\begin{figure*}[t!]
\begin{center}
\includegraphics[width=0.99\textwidth]{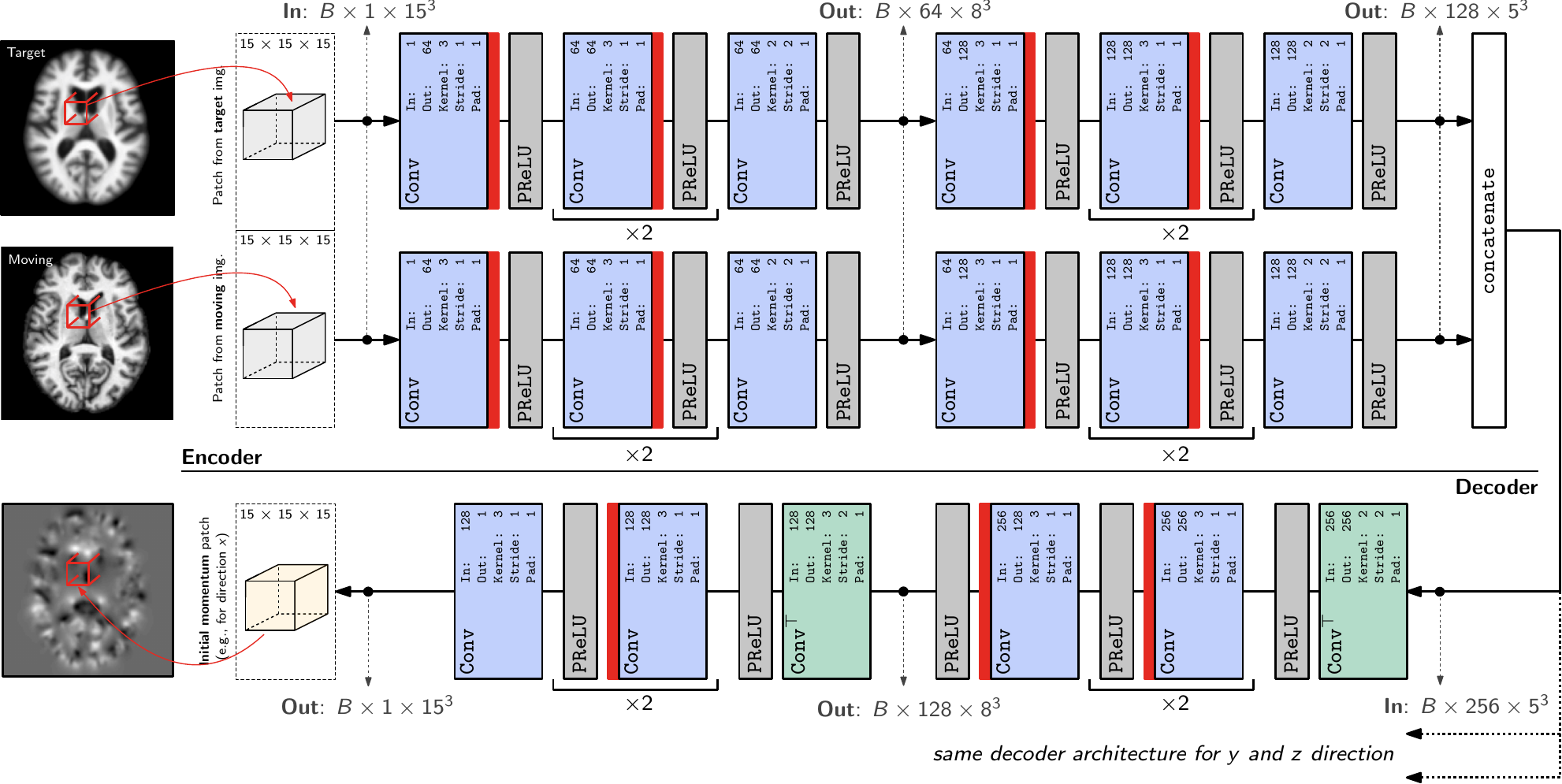}
\end{center}
\caption{3D (probabilistic) network architecture. The network takes two 3D patches from the moving and target image as the input, and outputs 3 3D initial momentum patches (one for each of the $x,y$ and $z$ dimensions respectively; for readability, only one decoder branch is shown in the figure). In case of the deterministic network, see Sec.~\ref{sec:deterministic}, the dropout layers, illustrated by \textcolor{red}{$\blacksquare$}, are removed. \xyrevisionsecond{\texttt{Conv}: 3D convolution layer. $\texttt{Conv}^{\intercal}$: 3D transposed convolution layer. Parameters for the \texttt{Conv} and $\texttt{Conv}^{\intercal}$ layers: \texttt{In}: input channel. \texttt{Out}: output channel. \texttt{Kernel}: 3D filter kernel size in each dimension. \texttt{Stride}: stride for the 3D convolution. \texttt{Pad}: zero-padding added to the boundaries of the input patch.} Note that in this illustration $B$ denotes the batch size.} 

\label{fig:network}
\end{figure*}

\subsection{Deep network for LDDMM prediction}
\label{sec:predict}
\xyrevision{The overall training strategy for our prediction models is as follows: We assume that we already have a set of LDDMM parameters which result in good registration results. We obtain these registration results by numerically optimizing the shooting formulation of LDMMM. These numerical optimizations can be based on images alone or could, of course, also make use of additional information available at training time, for example, object labels. For simplicity we only use image information here, but note that using additional information during training may result in increased prediction performance. The resulting initial momenta serve as training data. The goal is then to train a model to locally {\it predict} initial momenta from image patches of the moving and the target images. These predicted momenta should be good approximations of the initial momenta obtained via numerical optimization. In short, \emph{we train our deep learning framework to predict the initial momenta from image patches based on training data obtained from numerical optimization of the LDDMM shooting formulation.}} \xyrevisionsecond{During testing, we predict the initial momenta for the test image pairs, and generate the predicted deformation result simply by performing LDDMM shooting.}

Fig.~\ref{fig:network} shows the structure of the initial momentum prediction network. We first discuss the deterministic version of the network without dropout layers. We then introduce the Bayesian version of our network where dropout layers are used to convert the architecture into a probabilistic deep network. Finally, we discuss our strategy for patch pruning to reduce the number of patches needed for whole image prediction.

\xyrevisiondl{

\subsubsection{Deterministic network}
\label{sec:deterministic}
Our goal is to learn a \emph{prediction function} that 
takes two input patches, extracted at the same location\footnote{The locations of these patches are the same locations with respect to 
image grid coordinates, as the images are still unregistered at this point.} from the \emph{moving} and \emph{target image}, and predicts a desired  
initial vector-valued momentum patch, separated into the $x$, $y$ and $z$ dimensions, respectively. This prediction function should be learned from 
a set of training sample patches. These initial vector-valued momentum patches are obtained by numerical optimization of the LDDMM shooting formulation. 
More formally, given a 3D patch of size
$p\times p \times p$ voxels, we want to  
learn a function $f: \mathbb{R}^{3p} \times \mathbb{R}^{3p} \to \mathbb{R}^{9p}$.
In our formulation, $f$ is implemented by a deep neural network. 
Ideally, for two 3D \xyrevision{image} patches
$(\mathbf{u},\mathbf{v})=\mathbf{x}'$, with $\mathbf{u},\mathbf{v} \in \mathbb{R}^{3p}$, 
we want $\mathbf{y}' = f(\mathbf{x}')$ to be as close as possible to the desired \xyrevision{LDDMM optimization momentum patch}
$\mathbf{y}$ with respect to 
an appropriate loss function (e.g., the 1-norm). Our proposed architecture (for $f$) 
consists of two parts: an 
\emph{encoder} and a \emph{decoder} which we describe next.

\vskip0.5ex
\noindent
\textbf{Encoder.} The \textbf{Encoder} consists of two parallel encoders 
which learn features from the moving/target image patches 
independently. Each encoder contains two blocks of three $3 \times 3\times 3$ \xyrevision{3D} convolution \xyrevision{layers} and \xyrevision{PReLU~\cite{PReLU} activation} layers, followed
by another $2 \times 2\times 2$ convolution+PReLU with a stride of two, cf. Fig.~\ref{fig:network}. \xyrevision{The convolution layers with a stride of two reduce the size of the output patch, and essentially perform pooling operations.} \xyrevision{PReLU is an extension of the ReLU activation~\cite{ReLU}, \rknew{given as}
\[
\text{PReLU}(x) = 
\begin{cases}
x,\quad \text{if}\ x>0 \\
ax, \quad \text{otherwise}\enspace,
\end{cases}
\]
where $a$ is a parameter that is learned when training the network. In contrast to ReLU, PReLU avoids a zero gradient for negative inputs, effectively improving the network performance.} The number of features in the first block is 
$64$ and increases to $128$ in the second 
block. The learned features from the two encoders are then concatenated and sent to three parallel decoders (one \rknew{per} dimension $x,y,z$). 
}
\vskip0.5ex
\noindent
\textbf{Decoder.} Each decoder's structure is the inverse of the encoder, except 
that the number of features is doubled  ($256$ in the first block and $128$ in the second block) as the decoder's input is obtained from the {\it two} encoder branches. \xyrevision{We use 3D transposed convolution layers~\cite{Long2015} with a stride of 2, which are shown as the cyan layers in Fig.~\ref{fig:network} and can be regarded as the backward propagation of 3D convolution operations, to perform ``unpooling''.} We also omit the non-linearity after the 
final convolution layer, cf.~Fig.~\ref{fig:network}.

\vskip0.5ex
The idea of using convolution and \xyrevision{transpose of convolution} to learn the pooling/unpooling operation is motivated by~\cite{SpringenbergDBR14}, 
and it is especially suited for our network as the two encoders perform pooling independently which prevents us from using the pooling index for unpooling in the decoder. During training, we use the \emph{1-norm} between the predicted and the desired momentum to measure the prediction error. 
\xyrevision{We \rknew{chose} the 1-norm instead of the 2-norm as our loss function to be able to tolerate outliers} \xyrevisionsecond{and to generate sharper momentum predictions.} \mnminor{Ultimately, we are interested in predicting the deformation map and not the patch-wise momentum. However, this would require forming the entire momentum image from a collection of patches followed by shooting as part of the network training. Instead, predicting the momentum itself patch-wise significantly simplifies the network training procedure.} 
\mnr{\rknew{Also note that,} while we predict the momentum patch-by-patch, smoothing is performed over the full momentum image (reassembled from the patches) based on the smoothing kernel, $K$, of LDDMM. Specifically,} when predicting the deformation parameters for the whole image, we follow a sliding window strategy to predict the initial momentum in a patch-by-patch manner and then average the overlapping areas of the patches to obtain the final prediction result. 

\xyrevisionsecond{The number of 3D filters used in the network is 975,360. The overall number of parameters is 21,826,344. While this is a large number of parameters, we also have a very large number of training patches. For example, in our image-to-image registration experiments (see Sec.~\ref{sec:experiment}), the total number of $15\times15\times15$ 3D training patches 
\rknew{to train} the prediction network is 1,002,404. This amounts to approximately  3.4 billion voxels and is much larger than the total number of parameters in the network. Moreover, recent research~\cite{Gao2016} suggests that the degrees of freedom for a deep network can be significantly smaller than the number of its parameters.}

\vskip1ex
One question that naturally arises is why to use independent encoders/decoders in the prediction network. For the decoder part, we observed that an independent decoder structure is much easier to train than a network with one large decoder (\xyrevision{3 times} the number of features of a single decoder in our network) to predict the initial momentum in all dimensions simultaneously. In our experiments, such a combined network easily got stuck in poor local minima. As to the encoders, experiments do not show an obvious difference \rknew{in} prediction accuracy between using two independent encoders and one single large encoder. However, such a two-encoder strategy is beneficial when extending the approach to multi-modal image registration~\cite{yang2017multimodal}. Hence, using a two-encoder strategy here will make the approach easily retrainable for multi-modal image registration. In short, our network structure can be viewed as a multi-input multi-task network, where each encoder learns features for one patch source, and each decoder uses the shared image features from the encoders to predict one spatial dimension of the initial momenta. \mnr{
\rknew{We remark that, if one were} to predict the scalar-valued momentum, $\lambda$, instead of the vector-valued momentum, $m$, the network architecture could remain largely unchanged. The main difference would be that only one decoder would be required. Due to the simpler network architecture such an approach could potentially speed-up predictions. However, it remains to be investigated how such a network would perform in practice as the vector-valued momentum has been found to numerically better behave for LDDMM optimizations~\cite{Singh2013}.}

\subsubsection{Probabilistic network} 
\label{sec:probabilistic}

We extend our architecture to a probabilistic network using dropout~\cite{srivastava14a}, which can be viewed as (Bernoulli) approximate inference in Bayesian neural networks~\cite{Gal2015Bayesian,Gal16a}. \xyrevisiondl{In the following, we briefly review
the basic concepts, but refer the interested reader to the corresponding 
references for further technical details.

\vskip1ex
In our problem setting, we are given training patch tuples $\mathbf{x}_i = (\mathbf{u}_i,\mathbf{v}_i)$ with 
associated desired initial momentum patches $\mathbf{y}_i$. We denote the collection of this training
data by $\mathbf{X}$ and $\mathbf{Y}$. In the standard, non-probabilistic setting, we aim for predictions of the form 
$\mathbf{y}' = f(\mathbf{x}')$, given a new input patch $\mathbf{x}'$, where $f$ is
implemented by the proposed encoder-decoder network. In the \emph{probabilistic}
setting, however, the goal is to make predictions of the form 
$p(\mathbf{y}'|\mathbf{x}',\mathbf{X},\mathbf{Y})$. 
As this predictive distribution is intractable for most underlying models (as it
would require integrating over all possible models, and neural networks in
particular), the idea is to condition the model on a set of 
random variables $\mathbf{w}$. In case of (convolutional) 
neural networks with $N$ layers, these random variables are the 
weight matrices, i.e., $\mathbf{w} = (\mathbf{W}_i)_{i=1}^N$. 
However, evaluation of the predictive distribution 
$p(\mathbf{y}'|\mathbf{x}',\mathbf{X},\mathbf{Y})$ then requires 
the posterior 
over the weights  
$p(\mathbf{w}|\mathbf{X},\mathbf{Y})$ which can (usually) not be 
evaluated analytically. Therefore, in variational inference, 
$p(\mathbf{w}|\mathbf{X},\mathbf{Y})$ is 
replaced by a tractable variational distribution $q(\mathbf{w})$ and 
one minimizes the Kullback-Leibler divergence between 
$q(\mathbf{w})$ and $p(\mathbf{w}|\mathbf{X},{Y})$ with respect to 
the variational parameters $\mathbf{w}$.
This turns out to be equivalent to maximization of the \emph{log
evidence lower bound (ELBO)}. When the variational 
distribution is defined as
\begin{equation}
q(\mathbf{W}_i) = \mathbf{M}_i\cdot \text{diag}([z_{i, j}]_{j = 1}^{K_i}),\quad z_{i, j} \sim \text{Bernoulli}(d)\enspace,
\label{eqn:bernoullie}
\end{equation}
\xy{where $\mathbf{M}_i$ is the convolutional weight}, $i=1,\ldots,N$, $d$ is the probability that $z_{i, j} = 0$ and $K_i$ is chosen appropriately to match the dimensionality of $\mathbf{M}_i$,  Gal et al.~\cite{Gal2015Bayesian} show that ELBO 
maximization is achieved by training with dropout \cite{srivastava14a}. 
In the case of convolutional neural networks, dropout is applied after
each convolution layer (with dropout probability $d$)\footnote{with
additional $l_2$ regularization on the weight matrices of each layer.}.
In Eq.~\eqref{eqn:bernoullie}, $\mathbf{M}_i$ is the variational parameter 
which is optimized during training. 
Evaluation of the predictive distribution 
$p(\mathbf{y}'|\mathbf{x}',\mathbf{X},\mathbf{Y})$ can then be approximated 
via Monte-Carlo integration, i.e.,
\begin{equation}
p(\mathbf{y}'|\mathbf{x}',\mathbf{X},\mathbf{Y}) \approx \frac{1}{T}
\sum_{t=1}^T \hat{f}(\mathbf{x}', \hat{\mathbf{w}})\enspace.
\label{eqn:predictivedistributioneval}
\end{equation}
In detail, this corresponds to averaging the output of $T$ forward passes through the network with
dropout \emph{enabled}. Note that $\hat{f}$ and $\hat{\mathbf{w}}$ now 
correspond to random variables, as dropout means that we sample, in 
each forward pass, which connections are dropped.
In our implementation, we add dropout layers after all convolutional layers except for those used as pooling/unpooling layers (which are considered non-linearities applied to the
weight matrices \cite{Gal2015Bayesian}), \xyrevision{as well as the final convolution layer in the decoder, which generates the predicted momentum}. We train the network using stochastic gradient descent (SGD).

}
\vskip0.5ex
\noindent
\textbf{Network evaluation.} For testing, we keep the dropout layers \rknew{enabled} to maintain the probabilistic property of the network, and sample the network to obtain multiple momentum predictions for one moving/target image pair. We then choose the sample mean as the prediction result, see Eq.~\eqref{eqn:predictivedistributioneval}, and perform LDDMM shooting using all the samples to generate multiple deformation fields. The local variance of these deformation fields can then be used as an uncertainty estimate of the predicted deformation field. When selecting the dropout probability, \xyrevision{$d$}, a probability of $0.5$ would provide the largest variance, but may also enforce too much regularity for a convolutional network, especially in our case where dropout layers are added after every convolution layer. In our experiments, we use a dropout probability of $0.2$ (for all dropout units) as a balanced choice.

\begin{figure*}[h!]
\begin{center}
\includegraphics[width=0.90\textwidth]{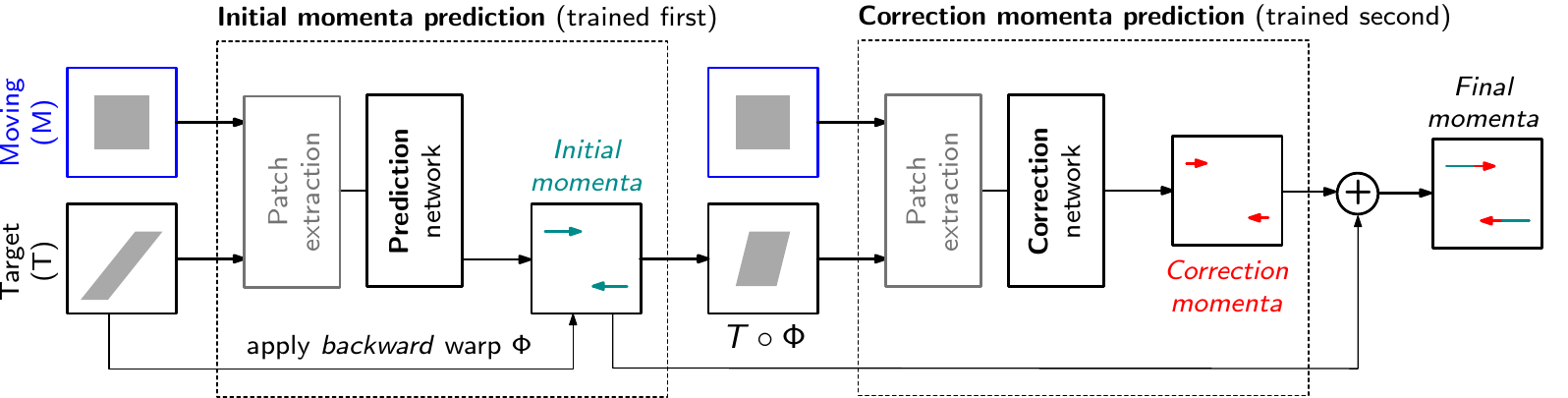}
\end{center}
\caption{The full prediction + correction architecture for LDDMM momenta. \mnr{First, a rough prediction of the initial momentum, $m_{LP}$, is obtained by the prediction network (LP) based on the patches from the unaligned moving image, $M$ and target image, $T$, respectively. The resulting deformation maps $\Phi^{-1}$ and $\Phi$ are computed by shooting. $\Phi$ is then applied to the target image to warp it to the space of the moving image. A second correction network is then applied to patches from the moving image $M$ and the warped target image $T\circ\Phi$ to predict a correction of the initial momentum, $m_C$ in the space of the moving image, $M$. The final momentum is then simply the sum of the predicted momenta, $m = m_{LP} + m_{C}$, which parameterizes a geodesic between the moving image and the target image.}}
\label{fig:correction_network}
\end{figure*}

\subsubsection{Patch pruning}
\label{sec:prune}

As discussed in Sec.~\ref{sec:deterministic}, we use a sliding-window approach to predict the deformation parameters (the momenta for \texttt{Quicksilver}) patch-by-patch for a whole image. Thus, computation time is proportional to the number of the patches we need to predict. When using a 1-voxel sliding window stride, the number of patches to predict for a whole image could be substantial. For a typical 3D image of size $128\times128\times128$ using a $15\times15\times15$ patch for prediction will require more than 1.4 million patch predictions. Hence, we use two techniques to drastically reduce the number of patches needed for deformation prediction. First, we perform patch pruning by ignoring all patches that belong to the background of both the moving image and the target image. This is justified, because according to LDDMM theory the initial momentum in constant image regions, and hence also in the image background, should be zero. Second, we use a large voxel stride (e.g., 14 for $15\times15\times15$ patches) for the sliding window operations. This is reasonable for our initial momentum parameterization because of the compact support (at edges) of the initial momentum and the spatial shift invariance we obtain via the pooling/unpooling operations. By using these two techniques, we can reduce the number of predicted patches for one single image dramatically. For example, by $99.995\%$ for 3D brain images of dimension $229\times 193 \times 193$.

\subsection{Correction network}
\label{sec:correct}
There are two main shortcomings of the deformation prediction network. (i) The complete iterative numerical approach typically used for LDDMM registration is replaced by a {\it single} prediction step. Hence, it is not possible to recover from any prediction errors. (ii) To facilitate training a network with a small number of images, to make predictions easily parallelizable, and to be able to perform predictions for large 3D image volumes, the prediction network predicts the initial momentum {\it patch-by-patch}. However, since patches are extracted at the same spatial grid locations from the moving and target images, large deformations may result in drastic appearance changes between a source and a target patch. In the extreme case, corresponding image information may no longer be found for a given source and target patch pair. This may happen, for example, when a small patch-size encounters a large deformation. While using larger patches would be an option (in the extreme case the entire image would be represented by one patch), this would require a network with substantially larger capacity (to store the information for larger image patches and all meaningful deformations) and would also likely require much larger training datasets\footnote{\mn{In fact, we have successfully trained prediction models with as little as ten images using all combinations of pair-wise registrations to create training data~\cite{yang2017multimodal}. This is possible, because even in such a case of severely limited training data the number of {\it patches} that can be used for training is very large.}}.

\vskip0.5ex
To address these shortcomings, we propose a two-step prediction approach to improve overall prediction accuracy. The first step is our already described prediction network. We refer to the second step as the \emph{correction network}. The task of the correction network is to compensate for prediction errors of the first prediction step. The idea is grounded in two observations: The first observation is that patch-based prediction is accurate when the deformation inside the patch is small. This is sensible as the initial momentum is concentrated along the edges, small deformations are commonly seen in training images, and less deformation results in less drastic momentum values. Hence, more accurate predictions are expected for smaller deformations. Our second observation is that, given the initial momentum, we are able to generate the whole geodesic path using the geodesic shooting equations. Hence, we can generate two deformation maps: the forward warp $\Phi^{-1}$ that maps the moving image to the coordinates of the target image, and the backward warp $\Phi$ mapping the target image back to the coordinates of the moving image. Hence, after the first prediction step using our prediction network, we can warp the target image back to the moving image $M$ via $T\circ\Phi$. We can then train the {\it correction network} based on the difference between the moving image $M$ and the warped-back target image $T\circ\Phi$, such that it makes adjustments to the initial momentum predicted in the first step by our prediction network. Because $M$ and $T\circ\Phi$ are in the same coordinate system, the differences between these two images are small as long as the predicted deformation is reasonable, and more accurate predictions can be expected. Furthermore, the correction for the initial momentum is then performed in the original coordinate space (of the moving image) which allows us to obtain an overall corrected initial momentum, $m_0$. This is for example a useful property when the goal is to do statistics with respect to a fixed coordinate system, for example, an atlas coordinate system.

Fig.~\ref{fig:correction_network} shows a graphical illustration of the resulting two-step prediction framework. In the framework, the correction network has the same structure as the prediction network, and the only difference is the input of the networks and the output they produce. 
\xyrevision{
Training the overall framework is done sequentially:
\begin{enumerate}
  \item Train the prediction network using training images and the ground truth initial momentum obtained by numerical optimization of the LDDMM registration model.
  \item Use the {\it predicted} momentum from the prediction network to generate deformation fields to warp the target images in the training dataset back to the space of the moving images.
  \item Use the moving images and the warped-back target images to train the correction network. The correction network learns to predict the \textsl{difference} between the ground truth momentum and the predicted momentum from the prediction network.
\end{enumerate}
Using the framework during testing is similar to the training procedure, except here the outputs from the prediction network (using moving and target images as input) and the correction network (using moving and warped-back target images as input) are summed up to obtain the final predicted initial momentum. This summation is justified from the LDDMM theory as it is performed in a fixed coordinate system (a fixed tangent space), which is the coordinate system of the moving image. Experiments show that our prediction+correction approach results in lower training and testing error compared with only using a prediction network, as shown in \xyrevisionsecond{Sec.~\ref{subsec:setup} and} Sec.~\ref{sec:experiment}.
}
\begin{figure*}[t!]
    \begin{center}
    \begin{subfigure}[t]{0.49\textwidth}
        \centering
        \includegraphics[width=\textwidth]{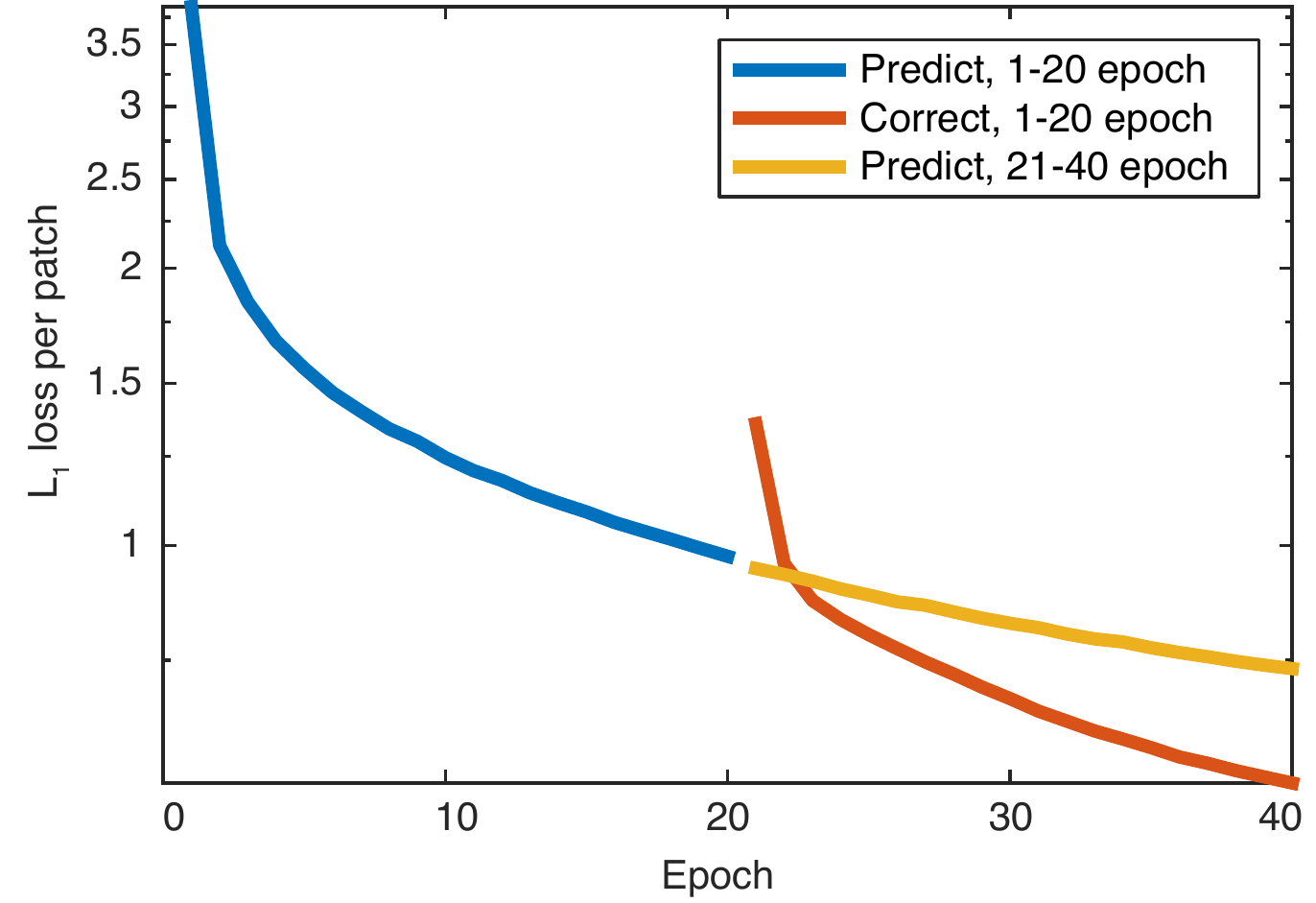}
        \caption{\texttt{Atlas-to-Image}}
    \end{subfigure}%
    ~ 
    \begin{subfigure}[t]{0.49\textwidth}
        \centering
        \includegraphics[width=0.99\textwidth]{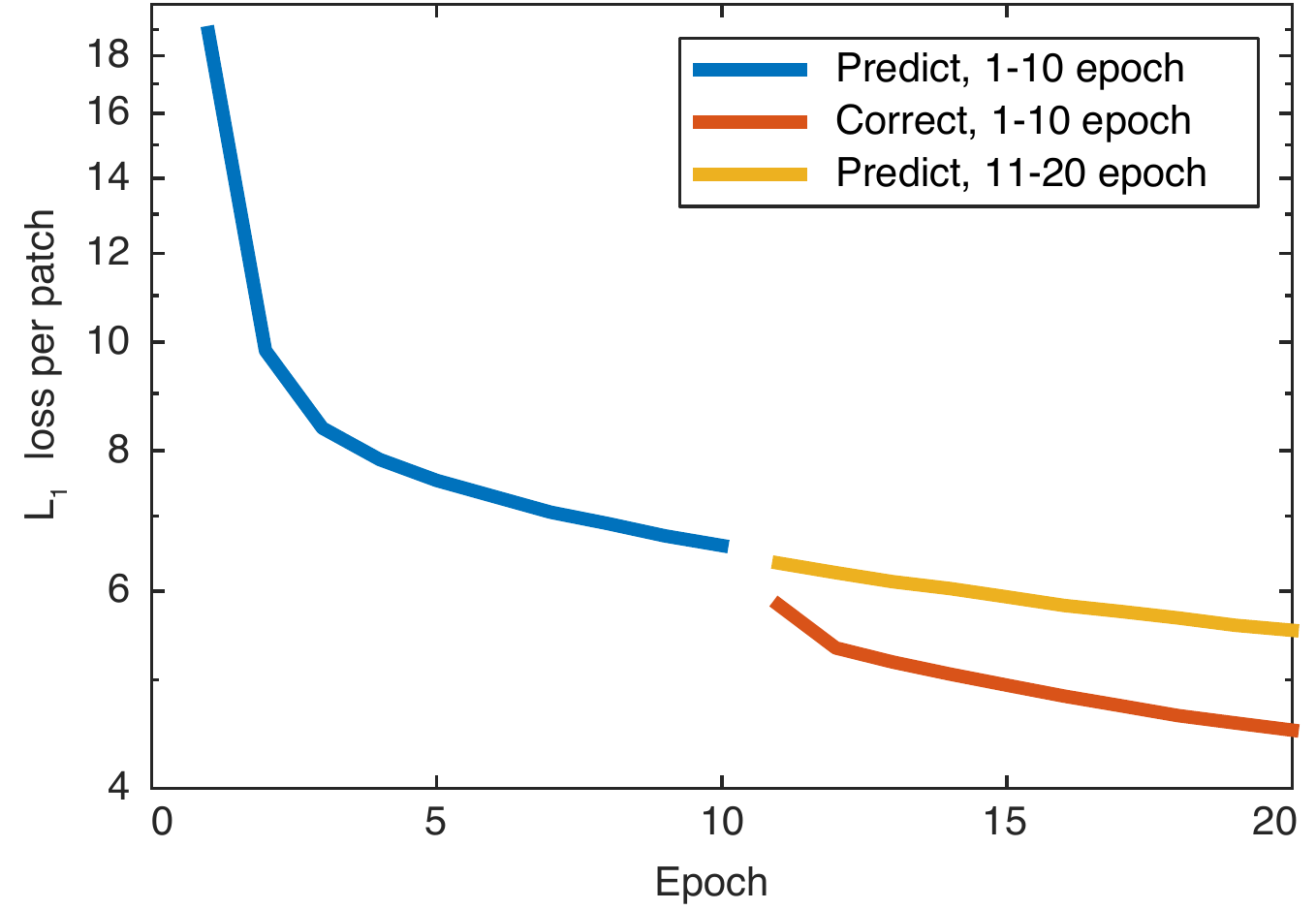}
        \caption{\texttt{Image-to-Image}}
    \end{subfigure}
    \end{center}
    \caption{$\text{Log}_{10}$ plot of $l_1$ training loss per patch. The loss is averaged across all iterations for every epoch for both the Atlas-to-Image case and the Image-to-Image case. \mnr{The combined prediction + correction networks obtain a lower loss per patch than the loss obtained by simply training the prediction networks for more epochs.}}
  	\label{fig:train_loss}
\end{figure*}
\subsection{Datasets / Setup}
\label{subsec:setup}
We evaluate our method using three 3D brain image registration experiments. 
The first experiment is designed to assess \emph{atlas-to-image} registration. In this experiment,  the moving image is always the atlas image. The second experiment addresses general \emph{image-to-image} registration. \xyrevision{The final experiment explores \emph{multi-modal image} registration; specifically, the registration of T1-weighted (T1w) and T2-weighted (T2w) magnetic resonance images.}

\begin{table*}[t]
\footnotesize
\begin{center}
\begin{tabular}{|r|c|c|c|c|c|c|c|c|}
\hline
& \multicolumn{7}{c|}{\textbf{Deformation Error for each voxel} [mm]} & \bm{$\text{\textbf{det}} J > 0$}\\ \hline\hline
\textsl{\xyrevision{Data percentile for all voxels}} & 0.3\% & 5\% & 25\% & 50\% & 75\% & 95\% & 99.7\% &\\ \hline
\texttt{Affine} & 0.0613	& 0.2520 & 0.6896 & 1.1911 & 1.8743 & 3.1413 & 5.3661 & N/A\\ \hline
\texttt{D}, velocity, stride 5 & 0.0237	& 0.0709 & 0.1601 & 0.2626 & 0.4117 & 0.7336 & 1.5166 & 100\%\\ \hline
\texttt{D}, velocity, stride 14 & 0.0254 & 0.075 & 0.1675 & 0.2703 & 0.415 & 0.743 & 1.5598 & 100\%\\ \hline
\texttt{D}, deformation, stride 5 & 0.0223 & 0.0665 & 0.1549 & 0.2614 &  0.4119 & 0.7388 & 1.5845 & 56\%\\ \hline
\texttt{D}, deformation, stride 14 & 0.0242 & 0.0721 & 0.1671 & 0.2772 & 0.4337 & 0.7932 & 1.6805 & 0\% \\ \hline\hline
\textbf{\texttt{P}, momentum, stride 14, 50 samples} & 0.0166 & 0.0479 & 0.1054  & 0.1678  & 0.2546 & 0.4537 &  1.1049 &  100\% \\ \hline
\textbf{\texttt{D}, momentum, stride 5} & 0.0129 & 0.0376 & 0.0884 & 0.1534 & 0.2506 & 0.4716 & 1.1095 & 100\%\\ \hline
\textbf{\texttt{D}, momentum, stride 14} & 0.013 & 0.0372 & 0.0834 & 0.1359 & 0.2112 & 0.3902 & 0.9433 & 100\%\\ \hline
\textbf{\texttt{D}, momentum, stride 14, 40 epochs} & 0.0119 & 0.0351 & 0.0793 & 0.1309 & 0.2070 & 0.3924 & 0.9542 & 100\%\\ \hline
\textbf{\texttt{D}, momentum, stride 14 + correction} & \cellcolor{green!30}\textbf{0.0104} & \cellcolor{green!30}\textbf{0.0309} & \cellcolor{green!30}\textbf{0.0704} & \cellcolor{green!30}\textbf{0.1167} & \cellcolor{green!30}\textbf{0.185} & \cellcolor{green!30}\textbf{0.3478} & \cellcolor{green!30}\textbf{0.841} & 100\%\\ \hline
\end{tabular}
\end{center}
\caption{Test result for \emph{atlas-to-image} registration. \xyrevisionsecond{\mnr{The table shows the distribution of the \rknew{2}-norm of the deformation error of the predicted deformation with respect to the deformation obtained by numerical optimization. Percentiles of the displacement errors are shown to provide a complete picture of the error distribution over just reporting the mean or median errors over all voxels within the brain mask in the dataset.}} \texttt{D}: deterministic network; \texttt{P}: probabilistic network; stride: stride length of the sliding window for whole image prediction; velocity: predicting initial velocity; deformation: predicting the deformation field; \xyrevision{momentum: predicting the initial momentum}; correction: using the correction network. The \bm{$\text{\textbf{det}} J > 0$} column shows the ratio of test cases with only positive-definite determinants of the Jacobian of the deformation map to the overall number of registrations (100\% indicates that all registration results were diffeomorphic). Our initial momentum networks are highlighted in \xyrevision{\textbf{bold}}. The best results are also highlighted in \textbf{bold}.}
\label{table:OASIS}
\end{table*}

\vskip1ex
For the \emph{atlas-to-image} registration experiment, we use 3D image volumes from the OASIS longitudinal dataset~\cite{OASIS}. Specifically, we use the first scan of all subjects, resulting in 150 brain images. We select the first 100 images as our training target images and the remaining 50 as our test target images. We create an unbiased atlas~\cite{joshi2004} from all training data using \texttt{PyCA}\footnote{\url{https://bitbucket.org/scicompanat/pyca}}~\cite{Singh2013, Singh2013HGM}, and use the atlas as the moving image. We use the LDDMM shooting algorithm to register the atlas image to all 150 OASIS images. The obtained initial momenta from the training data are used to train our network; the remaining momenta are used for validation.

\vskip1ex
For the \emph{image-to-image} registration experiment, we use all 373 images from the OASIS longitudinal dataset as the training data, and randomly select target images from different subjects for every image, creating 373 registrations for \mnr{the training of our prediction and correction networks}. For testing, we choose the four datasets (\texttt{LPBA40}, \texttt{IBSR18}, \texttt{MGH10}, \texttt{CUMC12}) evaluated in~\cite{Klein2009786}. We perform LDDMM shooting for all training registrations, and follow the evaluation procedure described in~\cite{Klein2009786} to perform pairwise registrations within all datasets, resulting in a total of 2168 registration (1560 from \texttt{LPBA40}, 306 from \texttt{IBSR18}, 90 from \texttt{MGH10}, 132 from \texttt{CUMC12}) test cases.

\vskip1ex
\xyrevision{For the \emph{multi-modal} registration experiment, we use the IBIS 3D Autism Brain image dataset~\cite{hazlett2017early}. This dataset contains 375 T1w/T2w brain images from 2 years old subjects. We select 359 of the images for training and use the remaining 16 images for testing. For training, we randomly select T1w-T1w image pairs and perform LDDMM shooting to generate the optimization momenta. \emph{We then train the prediction and correction networks to predict the momenta obtained from LDDMM T1w-T1w optimization using the image patches from the corresponding T1w moving image and T2w target image as network inputs}. For testing, we perform pair-wise T1w-T2w registrations for all 16 test images, resulting in 250 test cases. For comparison, we also train a T1w-T1w prediction+correction network that performs prediction on the T1w-T1w test cases. This network acts as the ``upper-bound'' of the potential performance of our multi-modal networks as it addresses the uni-modal registration case and hence operates on image pairs which have very similar appearance. Furthermore, to test prediction performance when using very limited training data, we also train a multi-modal prediction network and a multi-modal prediction+correction network using only 10 of the 365 training images which are randomly chosen for training. \rknew{In particular}, we perform pair-wise T1w-T1w registration on the 10 images, resulting in 90 registration pairs. We then use these 90 registration cases to train the multi-modal prediction networks.}

\vskip1ex
\xyrevision{For skull stripping, we use FreeSurfer~\cite{FreeSurfer} for the OASIS dataset and AutoSeg~\cite{AutoSeg} for the IBIS dataset. The 4 evaluation datasets for image-to-image experiment are already skull stripped as described in~\cite{Klein2009786}}. All images used in our experiments are first affinely registered to the ICBM MNI152 nonlinear atlas~\cite{Grabner2006} using \texttt{NiftyReg}\footnote{\url{https://cmiclab.cs.ucl.ac.uk/mmodat/niftyreg}} and intensity normalized \xy{via histogram equalization} prior to atlas building and LDDMM registration. All 3D volumes are of size $229\times193\times193$ except for the LPBA dataset ($229\times193\times229$), where we add additional blank image voxels for the atlas to keep the cerebellum structure. LDDMM registration is done using \texttt{PyCA}\footnote{\url{https://bitbucket.org/scicompanat/pyca}}~\cite{Singh2013} with SSD as the image similarity measure. We set the parameters for the regularizer of LDDMM\footnote{This regularizer is too weak to assure a diffeomorphic transformation based on the {\it sufficient} regularity conditions discussed in~\cite{beg2005}. For these conditions to hold in 3D, $L$ would need to be at least a differential operator of order 6. However, as long as the obtained velocity fields $v$ are finite over the unit interval, i.e., $\int_0^1 \|v\|_L^2~dt<\infty$ for an $L$ of at least order 6, we will obtain a diffeomorphic transform~\cite{dupuis1998}. In the discrete setting, this condition will be fulfilled for finite velocity fields. To side-step this issue, models based on Gaussian or multi-Gaussian kernels~\cite{bruveris2012} could also be used instead.} to $L = -a\nabla^2 - b\nabla(\nabla\cdot) + c$ as $[a, b, c] = [0.01, 0.01, 0.001]$, and $\sigma$ in Eqn.~\ref{eqn:momentum_energy} to 0.2. We use a $15\times15\times15$ patch size for deformation prediction in all cases, and use a sliding window with step-size 14 to extract patches for training. \xyrevision{The only exception is for the multi-modal network which is trained using only 10 images, where we choose a step-size of 10 to generate more training patches.} \mnminor{Note that using a stride of 14 during training means that we are in fact discarding available training patches to allow for reasonable network training times. However, we still retain a very large number of patches for training. To check that our number of patches for training is sufficient, we performed additional experiments for the image-to-image registration task using smaller strides when selecting training patches. Specifically, we doubled and tripled the training size for the prediction network. These experiments indicated that increasing the training data size further only results in marginal improvements, which are clearly outperformed by a combined prediction + correction strategy. Exploring alternative network structures, which may be able to utilize larger training datasets, is beyond the scope of this paper, but would be an interesting topic for future research.} 

The network is implemented in \texttt{PyTorch}\footnote{\url{https://github.com/pytorch/pytorch}}, and optimized using \texttt{Adam}~\cite{kingma2014adam}. We set the learning rate to 0.0001 and keep the remaining parameters at their default values. We train the prediction network for 10 epochs for the image-to-image registration experiment \xyrevision{and the multi-modal image registration experiment}, and 20 epochs for the atlas-to-image experiment. The correction networks are trained using the same number of epochs as their corresponding prediction networks. Fig.~\ref{fig:train_loss} shows the $l_1$ training loss per patch averaged for every epoch for the atlas-to-image and the image-to-image experiments. \xyadd{For both, using a correction network in conjunction with a prediction network results in lower training error compared with training the prediction network for more epochs.}

\section{\xyrevisionsecond{Results}}
\label{sec:experiment}
\subsection{Atlas-to-Image registration}
For the atlas-to-image registration experiment, we test two different sliding window strides for our patch-based prediction method: stride = 5 and stride = 14. We trained additional prediction networks predicting the initial velocity $v_0 = Km_0$ and the displacement field $\Phi(1)-\text{id}$ of LDDMM to show the effect of different deformation parameterizations on deformation prediction accuracy. We generate the predicted deformation map by integrating the shooting equation~\ref{eqn:forward} for the initial momentum and the initial velocity parameterization respectively. For the displacement parameterization we can directly read-off the map from the network output. We quantify the deformation errors per voxel using the voxel-wise two-norm of the deformation error with respect to the result obtained via numerical optimization for LDDMM using \texttt{PyCA}. Table~\ref{table:OASIS} shows the error percentiles over all voxels and test cases.

\begin{figure*}[t!]
\begin{center}
\includegraphics[width=1.0\textwidth]{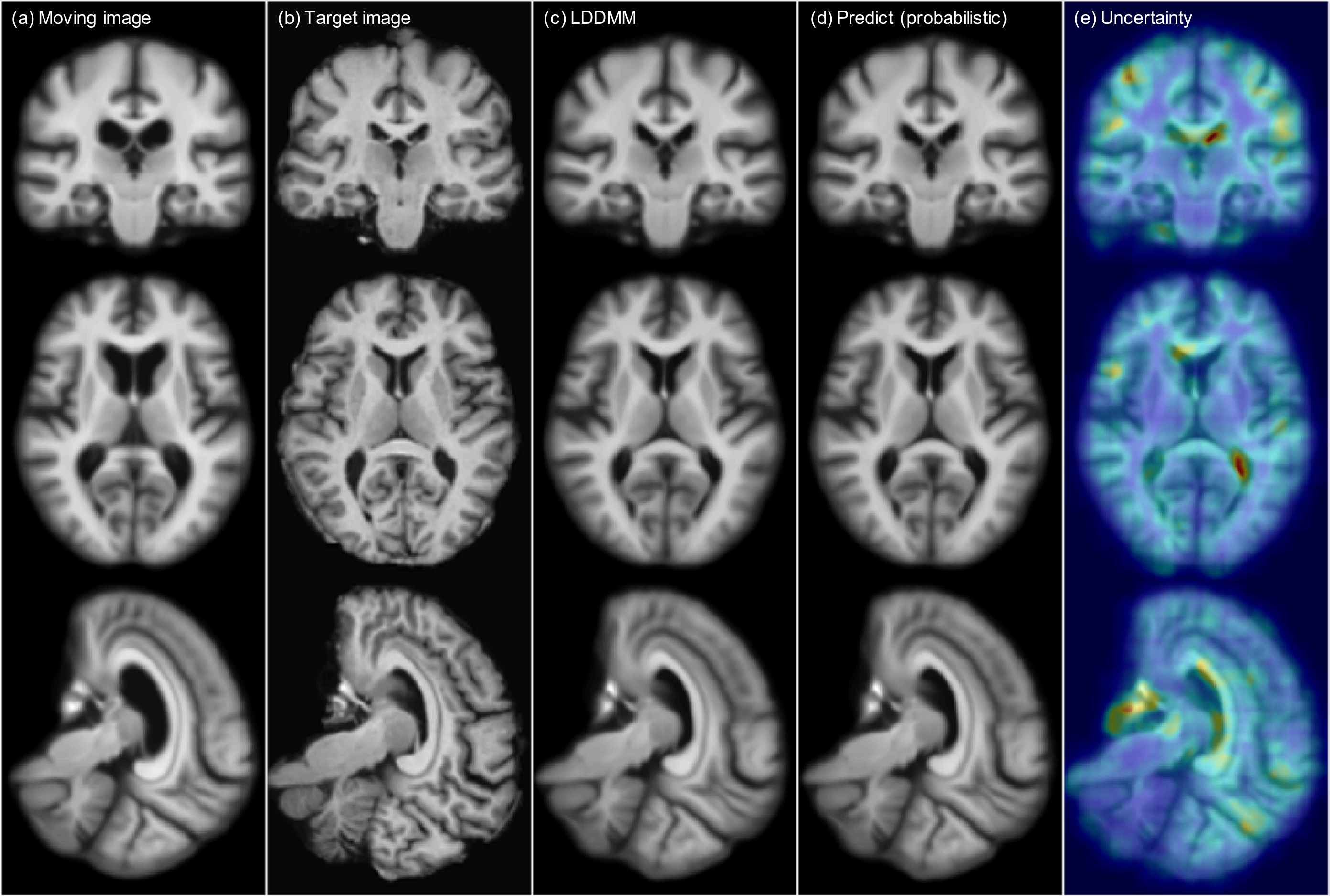}
\end{center}
\caption{Atlas-to-image registration example. From \emph{left} to \emph{right}: (a): moving (atlas) image; (b): target image; (c): deformation from optimizing LDDMM energy; (d): deformation from using the mean of 50 samples from the probabilistic network with stride=14 and patch pruning; (e): the uncertainty map as square root of the sum of the variances of the deformation in $x$, $y$, and $z$ directions mapped onto the predicted deformation result. The coloring indicates the level of uncertainty, with \textcolor{red}{red = high uncertainty} and \textcolor{blue}{blue = low uncertainty}. Best-viewed in color.}
\label{fig:OASIS_example}
\end{figure*}

We observe that the initial momentum network has better prediction accuracy compared to the results obtained via the initial velocity and displacement parameterization in both the 5-stride and 14-stride cases. This validates our hypothesis that momentum-based LDDMM is better suited for patch-wise deformation prediction. \xy{We also observe that the momentum prediction result using a smaller sliding window stride is slightly worse than the one using a stride of 14. This is likely the case, because in the atlas-to-image setting, the number of atlas patches that extract features from the atlas image is very limited, and using a stride of 14 during the training phase further reduces the available data from the atlas image. \xyadd{Thus, during testing, the encoder will perform very well for the 14-stride test cases since it has already seen all the input atlas patches during training. For a stride of 5 however, unseen atlas patches will be input to the network, resulting in reduced registration accuracy\footnote{\mn{This behavior could likely be avoided by randomly sampling patch locations during training instead of using a regular grid. However, since we aim at reducing the number of predicted patches we did not explore this option and instead maintained the regular grid sampling.}}.}} \mn{In contrast, the velocity and the displacement parameterizations result in slightly better predictions for smaller sliding window strides. That this is not the case for the momentum parameterization suggests that it is easier for the network to learn to predict the momentum, as it indeed has become more specialized to the training data which was obtained with a stride of 14.} One of the important properties of LDDMM shooting is its ability to generate diffeomorphic deformations. To assess this property, we calculate the local Jacobians of the resulting deformation maps. Assuming no flips of the entire coordinate system, a diffeomorphic deformation map should have positive Jacobian determinants everywhere, otherwise foldings occur in the deformation maps. We calculate the ratio of test cases with positive Jacobian determinants of the deformation maps to all test cases, shown as $\text{\textbf{det}} J > 0$ in Table~\ref{table:OASIS}. We observe that the initial momentum and the initial velocity networks indeed generate diffeomorphic deformations in all scenarios. However, the deformation accuracy is significantly worse for the initial velocity network. Predicting the displacement directly cannot guarantee diffeomorphic deformations even for a small stride. This is unsurprising as, similar to existing optical flow approaches~\cite{deepflow, flownet}, directly predicting displacements does not encode deformation smoothness. Hence, the initial momentum parameterization is the preferred choice among our three tested parameterizations as it achieves the best prediction accuracy and guarantees diffeomorphic deformations. Furthermore, the initial momentum prediction including the correction network with a stride of 14 achieves the best registration accuracy overall among the tested methods, even outperforming the prediction network alone trained with more training iterations (\textbf{\texttt{D}, stride 14, 40 epochs}). This demonstrates that the correction network is capable of improving the initial momentum prediction beyond the capabilities of the original prediction network.

Fig.~\ref{fig:OASIS_example} shows one example atlas-to-image registration case. The predicted deformation result is very similar to the deformation from LDDMM optimization. We compute the square root of the sum of the variance of the deformation in the $x$, $y$ and $z$ directions to quantify deformation uncertainty, and visualize it on the rightmost column of the figure. The uncertainty map shows high uncertainty along the ventricle areas where drastic deformations occur, as shown in the moving and target images.
\begin{figure*}[t!]
    \begin{center}
    \begin{subfigure}[t]{0.481\textwidth}
        \centering
        \includegraphics[width=\textwidth]{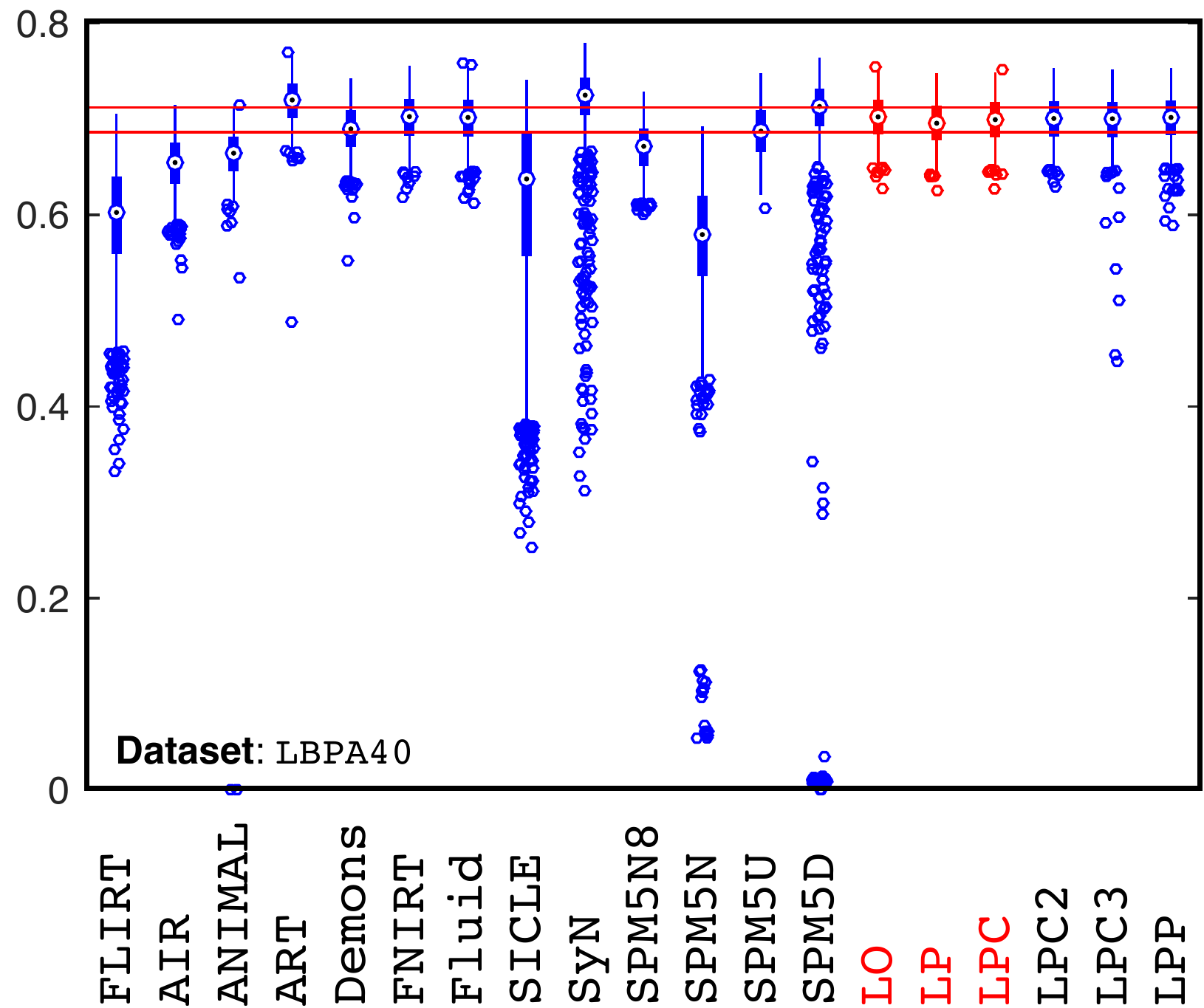}
    \end{subfigure}%
    ~ 
    \begin{subfigure}[t]{0.49\textwidth}
        \centering
        \includegraphics[width=0.99\textwidth]{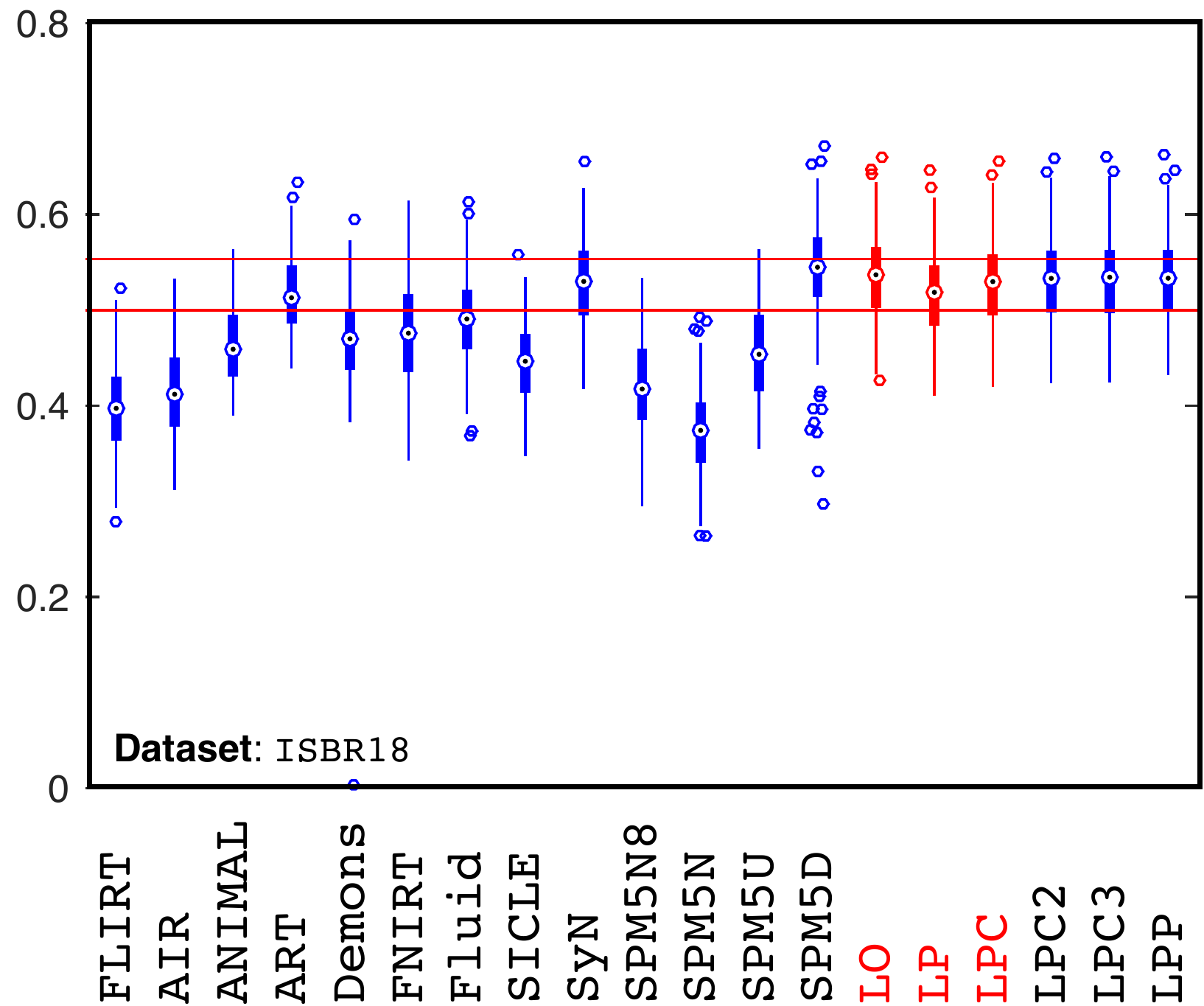}
    \end{subfigure}\\
    \vskip1.5ex
    \begin{subfigure}[t]{0.49\textwidth}
        \centering
        \includegraphics[width=0.977\textwidth]{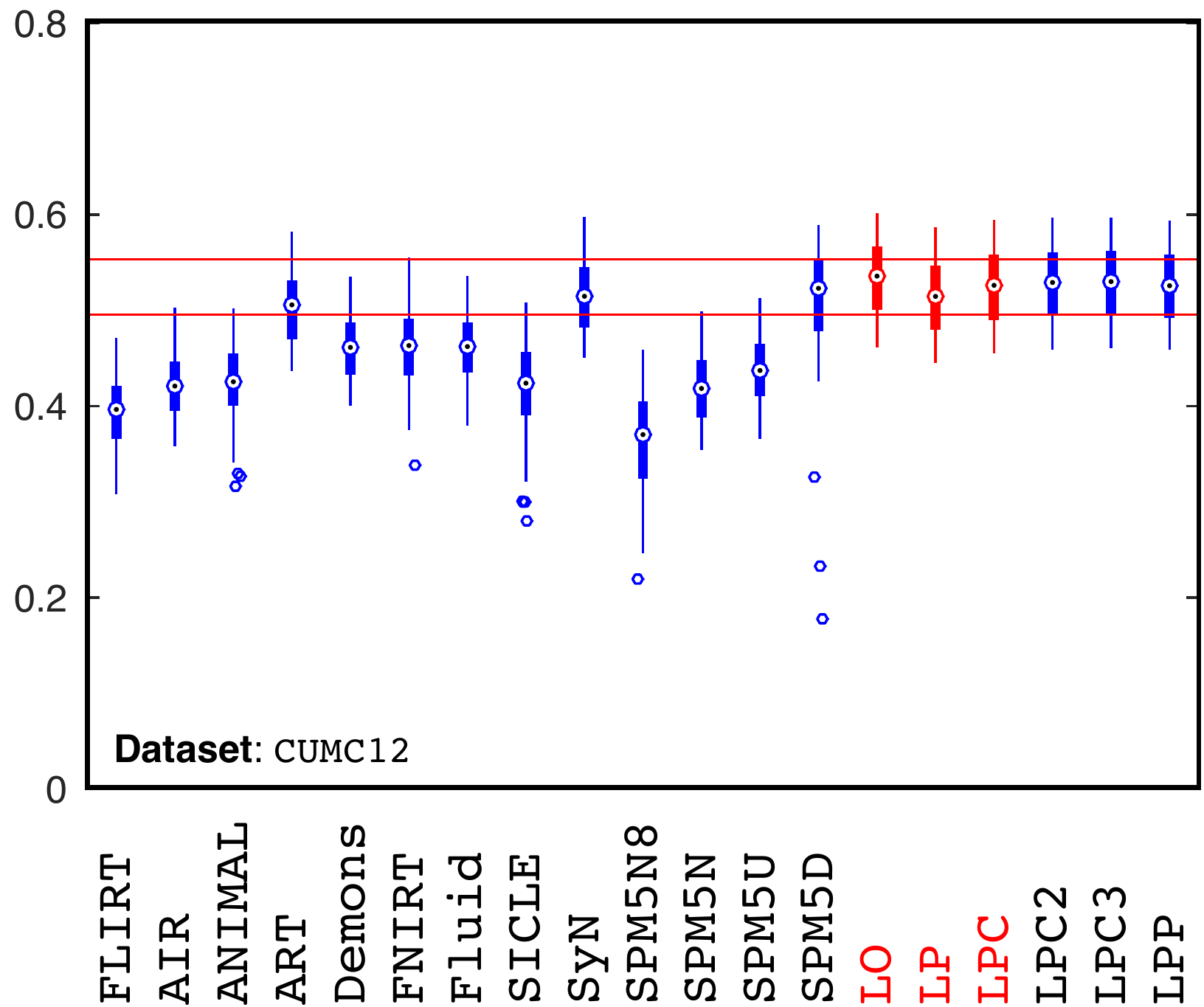}
    \end{subfigure}%
    ~ 
    \begin{subfigure}[t]{0.481\textwidth}
        \centering
        \includegraphics[width=0.997\textwidth]{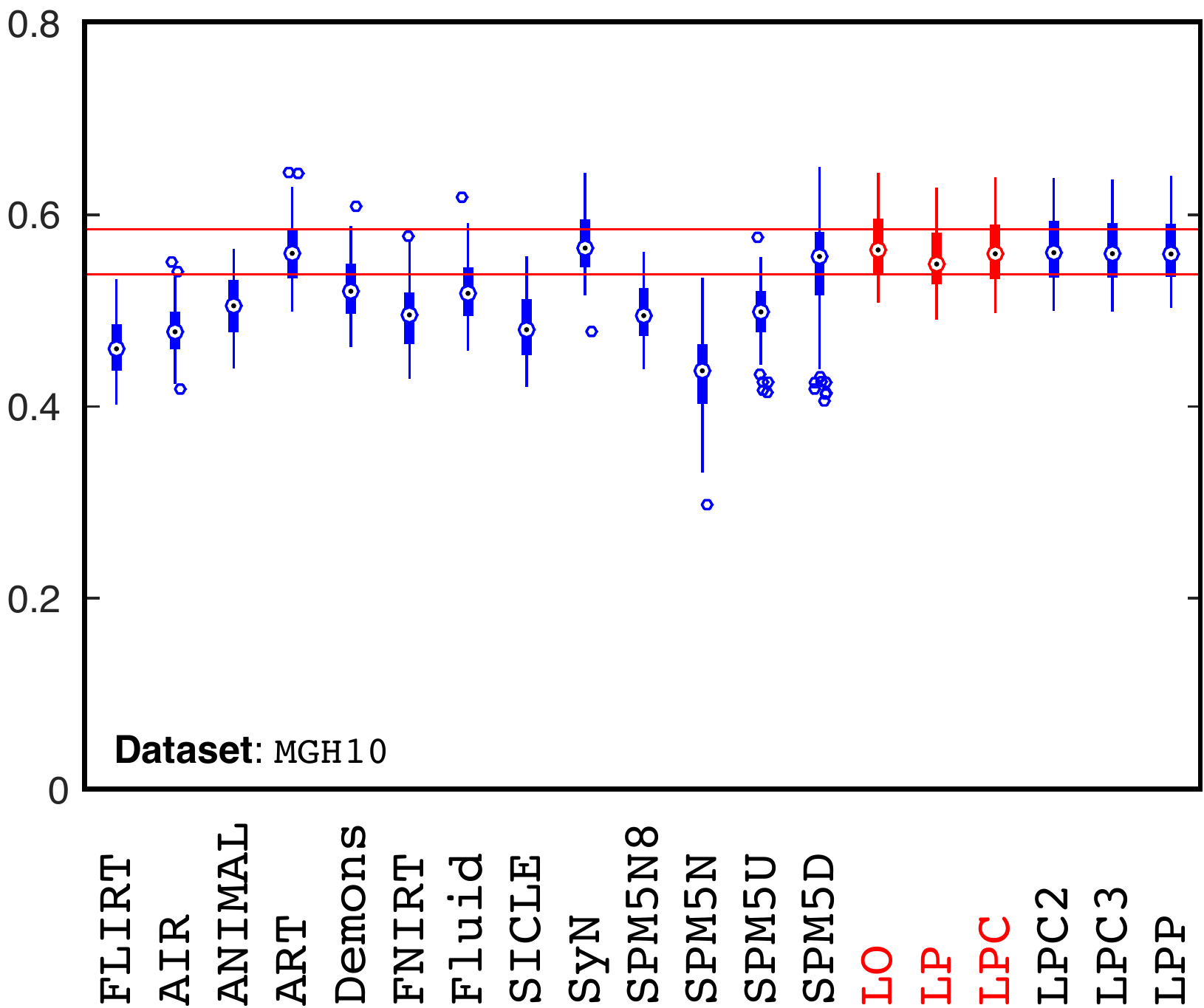}
    \end{subfigure}
    \end{center}
    \caption{Overlap by registration method for the \emph{image-to-image} registration case. The boxplots illustrate the mean target overlap measures averaged over all subjects in each label set, where mean target overlap is the average of the fraction of the target region overlapping with the registered moving region over all labels. \xyrevision{The proposed LDDMM-based methods in this paper} are highlighted in \textcolor{red}{red}. \texttt{LO} = LDDMM optimization; \texttt{LP} = prediction network; \texttt{LPC} = prediction network + correction network. \xyrevision{\texttt{LPP}: prediction network + using the prediction network for correction. \texttt{LPC2/LPC3}: prediction network + iteratively using the correction network 2/3 times.} Horizontal red lines show the \texttt{LPC} performance in the lower quartile to upper quartile (best-viewed in color). \xyrevisionsecond{The medians of the overlapping scores for [\texttt{LPBA40}, \texttt{IBSR18}, \texttt{CUMC12}, \texttt{MGH10}] for \texttt{LO}, \texttt{LP} and \texttt{LPC} are: \texttt{LO}: [0.702, 0.537, 0.536, 0.563]; \texttt{LP}: [0.696, 0.518, 0.515, 0.549]; \texttt{LPC}: [0.702, 0.533, 0.526, 0.559]. Best-viewed in color.}}
    \label{fig:image_image}
\end{figure*}

\subsection{Image-to-Image registration}
\xyadd{In this experiment, we use a sliding window stride of 14 for both the prediction network and the correction network during evaluation.} \xyrevision{We mainly compare the following three LDDMM-based -methods: (i) the numerical LDDMM optimization approach (\texttt{LO}) as implemented in \texttt{PyCA}, which acts as an upper bound on the performance of our prediction methods; and two flavors of \texttt{Quicksilver}: (ii) only the prediction network (\texttt{LP}) and (iii) the prediction+correction network (\texttt{LPC}). Example registration cases are shown in Fig. 9.}
\xyrevision{
\subsubsection{LDDMM energy}
\label{subsubsec:energy}
\begin{table}[htbp]
\footnotesize
\begin{center}
\begin{tabular}{|c|c|c|c|}
\hline
\multicolumn{4}{|c|}{\textbf{LDDMM energy} for image-to-image test datasets}\\ \hline\hline
\multicolumn{4}{|c|}{\texttt{LPBA40}}\\ \hline
\texttt{initial} & \texttt{LO} & \texttt{LP} & \texttt{LPC}\\\hline
$0.120\pm0.013$ & $0.027\pm0.004$ & $0.036\pm0.005$ & $0.030\pm0.005$\\\hline\hline
\multicolumn{4}{|c|}{\texttt{IBSR18}}\\ \hline
\texttt{initial} & \texttt{LO} & \texttt{LP} & \texttt{LPC}\\\hline
$0.214\pm0.032$ & $0.037\pm0.008$ & $0.058 \pm 0.013$ & $0.047 \pm 0.011$\\\hline\hline
\multicolumn{4}{|c|}{\texttt{CUMC12}}\\ \hline
\texttt{initial} & \texttt{LO} & \texttt{LP} & \texttt{LPC}\\\hline
$0.246\pm0.015$ & $0.044\pm 0.003$ & $0.071\pm 0.004$ & $0.056\pm 0.004$\\\hline\hline
\multicolumn{4}{|c|}{\texttt{MGH10}}\\ \hline
\texttt{initial} & \texttt{LO} & \texttt{LP} & \texttt{LPC}\\\hline
$0.217\pm 0.012$ & $0.039\pm 0.003$ & $0.062 \pm 0.004$ & $0.049\pm 0.003$\\\hline
\end{tabular}
\vskip 2ex
\caption{\label{table:energy}\xyrevision{Mean and standard deviation of the LDDMM energy for four image-to-image test datasets. \texttt{initial}: the initial LDDMM energy between the original moving image and the target image \mnr{after affine registration to the atlas space}, i.e. the original image matching energy. \texttt{LO}: LDDMM optimization. \texttt{LP}: prediction network. \texttt{LPC}: prediction+correction network.}}
\end{center}
\end{table}

To test the ability of our prediction networks to replace numerical optimization, we compare the LDDMM energies obtained using optimization from \texttt{LO} with the energies corresponding to the predicted momenta from \texttt{LP} and \texttt{LPC}. Low energies for the predicted momenta, which are comparable to the energies obtained by numerical optimization (\texttt{LO}), would suggest that our prediction models can indeed act as replacements for numerical optimization. \mnr{\rknew{However, note that, in general,} a low energy will only imply a good registration result if the registration model is fully appropriate for the registration task. Ultimately, registration quality should be assessed based on a particular task: most directly by measuring landmark errors or (slightly more indirectly) by measuring overlaps of corresponding regions as done in Section~\ref{subsubsec:overlap}.} Note also that our networks for image-to-image registration are trained on the OASIS dataset. Hence, improved results may be achievable by training dataset specific models. Table~\ref{table:energy} shows the results for the four test datasets. Compared with the initial LDDMM energy \mnr{based on affine registration to the atlas space} in the \texttt{initial} column, both \texttt{LP} and \texttt{LPC} have drastically lower LDDMM energy values; \rknew{further, these values} are only slightly higher than those for \texttt{LO}. Furthermore, compared with \texttt{LP}, \texttt{LPC} generates LDDMM energy values that are closer to \texttt{LO}, which indicates that using the prediction+correction approach \mnr{results in momenta which are closer to the optimal solution than the ones obtained by using the prediction network only.}
}

\subsubsection{Label overlap}
\label{subsubsec:overlap}
For image-to-image registration we follow the approach in \cite{Klein2009786} and calculate the target overlap (TO) of labeled brain regions after registration: $TO = \frac{|l_{m}\cap l_{t}|}{|l_{t}|}$, where $l_m$ and $l_t$ indicate the corresponding labels for the moving image (after registration) and the target image. We then evaluate the mean of the target overlap averaged first across all labels for every registration case. The evaluation results for other methods tested in \cite{Klein2009786} are available online. We compare our registration approaches to these results. \xyrevision{An interesting question is if the prediction network and the correction network are identical, and whether the prediction network can be used in the correction step. Another question is if the correction network can be applied multiple times in the correction step to further improve results. Thus, to test the usefulness of the correction network in greater depth, we also create three additional formulations of our prediction framework: (i) prediction network + using the same prediction network to replace the correction network in the correction step (\texttt{LPP}); (ii) applying the correction network twice (\texttt{LPC2}) and (iii) applying the correction network three times (\texttt{LPC3}).}

Fig.~\ref{fig:image_image} shows the evaluation results. Several points should be noted: first, the LDDMM optimization performance is on par with \texttt{SyN}~\cite{Avants200826}, \texttt{ART}~\cite{Ardekani200567} and the SPM5 DARTEL Toolbox (\texttt{SPM5D})~\cite{Ashburner200795}. This is reasonable as these methods are all non-parametric diffeomorphic or homeomorphic registration methods, allowing the modeling of large deformations between image pairs. Second, using only the prediction network results in a slight performance drop compared to the numerical optimization results (\texttt{LO}), but the result is still competitive with the top-performing registration methods. Furthermore, also using the correction network boosts the deformation accuracy nearly to the same level as the LDDMM optimization approach (\texttt{LO}). The red horizontal lines in Fig.~\ref{fig:image_image} show the lower and upper quartiles of the target overlap score of the prediction+correction method. Compared with other methods, our prediction+correction network achieves top-tier performance for label matching accuracy at a small fraction of the computational cost. Lastly, in contrast to many of the other methods \texttt{Quicksilver} produces virtually no outliers. One can speculate that this may be the benefit of learning to predict deformations from a \rknew{large} {\it population} of data, which may result in a prediction model which conservatively rejects unusual deformations. \mnminor{Note that such a population-based approach is very different from most existing registration methods which constrain deformations based on a regularizer chosen for a mathematical registration model. Ultimately, a deformation model for image registration should model what deformations are expected. Our population-based approach is a step in this direction, but, of course, still depends on a chosen regularizer to generate training data. Ideally, this regularizer itself should be learned from data.}

\xyrevision{An interesting discovery is that \texttt{LPP}, \texttt{LPC2} and \texttt{LPC3} produce label overlapping scores that are on-par with \texttt{LPC}. However, as we will show in Sec.~\ref{subsubsec:correct}, \texttt{LPP}, \texttt{LPC2} and \texttt{LPC3} deviate from our goal of predicting deformations that are similar to the LDDMM optimization result (\texttt{LO}). In fact, they produce more drastic deformations that can lead to worse label overlap and even numerical stability problems. These problems can be observed in the \texttt{LPBA40} results shown in Fig.~\ref{fig:image_image}, which show more outliers with low overlapping scores for \texttt{LPP} and \texttt{LPC3}. In fact, there are 12 cases for \texttt{LPP} where the predicted momentum cannot generate deformation fields via LDDMM shooting using \texttt{PyCA}, due to problems \rknew{related to} numerical integration. These cases are therefore not included in Fig.~\ref{fig:image_image}. \xyrevisionsecond{\texttt{PyCA} uses an explicit Runge-Kutta method (RK4) for time-integration. Hence, numerical instability is likely due to the use of a fixed step size for this time-integration which is small enough for the deformations expected to occur for these brain registration tasks, but which may be too large for the more extreme momenta \texttt{LPP} and \texttt{LPC3} create for some of these cases. Using a smaller step-size would regain numerical stability in this case.} 
}

\begin{table*}[t!]
\centering
\fontsize{6}{7}\selectfont
\begin{subtable}{1\linewidth}
\begin{center}
\begin{tabularx}{1\textwidth}{|*{9}{>{\hsize=0.11\linewidth}X|}}
\hline
\multicolumn{9}{|c|}{\textbf{Dataset}: \texttt{LPBA40}} \\ \hline\hline
 & \texttt{FLIRT} & \texttt{AIR} & \texttt{ANIMAL} & \texttt{ART} & \texttt{Demons} & \texttt{FNIRT} & \texttt{Fluid} & \texttt{SLICE} \\ \hline
\rowcolor{green!30} \cellcolor{white!30}\texttt{LO} & $0.108\pm0.054$ & $0.049\pm0.021$ & $0.039\pm0.029$ & \cellcolor{red!30}$-0.017\pm0.013$ & $0.012\pm0.014$ & $\cellcolor{blue!30}0.001\pm0.014$ & $0.001\pm0.013$ & $0.097\pm0.1$ \\ \hline
\rowcolor{green!30} \cellcolor{white!30}\texttt{LP} & $0.102\pm0.054$ & $0.043\pm0.02$ & $0.033\pm0.029$ & \cellcolor{red!30}$-0.024\pm0.013$ & $0.006\pm0.014$ & \cellcolor{red!30}$-0.006\pm0.014$ & \cellcolor{red!30}$-0.005\pm0.013$ & $0.091\pm0.1$\\ \hline
\rowcolor{green!30} \cellcolor{white!30}\texttt{LPC} & $0.106\pm0.054$ & $0.046\pm0.021$ & $0.037\pm0.029$ & \cellcolor{red!30}$-0.02\pm0.013$ & $0.009\pm0.014$ & \cellcolor{red!30}$-0.002\pm0.014$ & \cellcolor{red!30}$-0.002\pm0.013$ & $0.095\pm0.1$\\ \hline

 & \texttt{SyN} & \texttt{SPM5N8} & \texttt{SPM5N} & \texttt{SPM5U} & \texttt{SPM5D} & \texttt{LO} & \texttt{LP} & \texttt{LPC} \\ \hline
\rowcolor{green!30} \cellcolor{white!30}\texttt{LO}  & \cellcolor{red!30}$-0.013\pm0.05$ & $0.032\pm0.018$ & $0.13\pm0.07$ & $0.015\pm0.017$ & $0.03\pm0.16$ & \cellcolor{white!30}N/A  & $0.006\pm0.003$ & $0.003\pm0.002$\\ \hline
\rowcolor{green!30} \cellcolor{white!30}\texttt{LP}  & \cellcolor{red!30}$-0.02\pm0.05$ & $0.025\pm0.018$ & $0.124\pm0.07$ & $0.009\pm0.017$ & $0.023\pm0.16$ & \cellcolor{red!30}$-0.006\pm0.003$ &  \cellcolor{white!30}N/A  & \cellcolor{red!30}$-0.004\pm0.002$\\ \hline
\rowcolor{green!30} \cellcolor{white!30}\texttt{LPC}  & \cellcolor{red!30}$-0.016\pm0.05$ & $0.029\pm0.018$ & $0.127\pm0.07$ & $0.012\pm0.017$ & $0.027\pm0.16$ & \cellcolor{red!30}$-0.003\pm0.002$ & $0.004\pm0.002$ & \cellcolor{white!30}N/A\\ \hline
\end{tabularx}
\end{center}
\vskip2ex
\end{subtable}

\begin{subtable}{1\linewidth}
\begin{center}
\begin{tabularx}{1\textwidth}{|*{9}{>{\hsize=0.11\linewidth}X|}}
\hline
\multicolumn{9}{|c|}{\textbf{Dataset}: \texttt{IBSR18}} \\ \hline\hline
 & \texttt{FLIRT} & \texttt{AIR} & \texttt{ANIMAL} & \texttt{ART} & \texttt{Demons} & \texttt{FNIRT} & \texttt{Fluid} & \texttt{SLICE} \\ \hline
\rowcolor{green!30} \cellcolor{white!30}\texttt{LO} & $0.136\pm0.025$ & $0.119\pm0.03$ & $0.07\pm0.027$ & $0.018\pm0.022$ & $0.064\pm0.034$ & $0.057\pm0.026$ & $0.044\pm0.019$ & $0.088\pm0.029$\\ \hline
\rowcolor{green!30} \cellcolor{white!30}\texttt{LP} & $0.118\pm0.022$ & $0.101\pm0.028$ & $0.052\pm0.025$ & \cellcolor{blue!30}$0\pm0.021$ & $0.047\pm0.032$ & $0.039\pm0.023$ & $0.026\pm0.018$ & $0.07\pm0.027$\\ \hline
\rowcolor{green!30} \cellcolor{white!30}\texttt{LPC} & $0.129\pm0.024$ & $0.112\pm0.03$ & $0.063\pm0.027$ & $0.01\pm0.022$ & $0.058\pm0.033$ & $0.049\pm0.025$ & $0.036\pm0.019$ & $0.08\pm0.029$\\ \hline

 & \texttt{SyN} & \texttt{SPM5N8} & \texttt{SPM5N} & \texttt{SPM5U} & \texttt{SPM5D} & \texttt{LO} & \texttt{LP} & \texttt{LPC} \\ \hline
\rowcolor{green!30} \cellcolor{white!30}\texttt{LO} & $\cellcolor{blue!30}0.005\pm0.024$ & $0.112\pm0.034$ & $0.161\pm0.042$ & $0.08\pm0.030$ & \cellcolor{red!30}$-0.009\pm0.035$& \cellcolor{white!30}N/A & $0.018\pm0.007$ & $0.007\pm0.004$\\ \hline
\rowcolor{green!30} \cellcolor{white!30}\texttt{LP} & \cellcolor{red!30}$-0.013\pm0.024$ & $0.094\pm0.032$ & $0.144\pm0.042$ & $0.062\pm0.027$ & \cellcolor{red!30}$-0.026\pm0.035$ & \cellcolor{red!30}$-0.018\pm0.007$& \cellcolor{white!30}N/A & \cellcolor{red!30}$-0.01\pm0.004$\\ \hline
\rowcolor{green!30} \cellcolor{white!30}\texttt{LPC} & \cellcolor{blue!30}$-0.002\pm0.024$ & $0.105\pm0.034$ & $0.154\pm0.043$ & $0.073\pm0.029$ & \cellcolor{red!30}$-0.016\pm0.035$ & \cellcolor{red!30}$-0.007\pm0.004$ & $0.01\pm0.004$& \cellcolor{white!30}N/A\\ \hline

\end{tabularx}
\end{center}
\vskip2ex
\end{subtable}

\begin{subtable}{1\linewidth}
\begin{center}
\begin{tabularx}{1\textwidth}{|*{9}{>{\hsize=0.11\linewidth}X|}}
\hline
\multicolumn{9}{|c|}{\textbf{Dataset}: \texttt{CUMC12}} \\ \hline\hline
 & \texttt{FLIRT} & \texttt{AIR} & \texttt{ANIMAL} & \texttt{ART} & \texttt{Demons} & \texttt{FNIRT} & \texttt{Fluid} & \texttt{SLICE} \\ \hline
\rowcolor{green!30} \cellcolor{white!30}\texttt{LO}&$0.14\pm0.02$&$0.111\pm0.019$&$0.108\pm0.031$&$0.031\pm0.01$&$0.072\pm0.012$&$0.071\pm0.019$&$0.073\pm0.017$&$0.115\pm0.03$\\ \hline
\rowcolor{green!30} \cellcolor{white!30}\texttt{LP}&$0.12\pm0.017$&$0.092\pm0.017$&$0.089\pm0.031$&$0.012\pm0.01$&$0.052\pm0.01$&$0.052\pm0.017$&$0.053\pm0.015$&$0.096\pm0.031$\\ \hline
\rowcolor{green!30} \cellcolor{white!30}\texttt{LPC}&$0.131\pm0.018$&$0.102\pm0.018$&$0.1\pm0.031$&$0.023\pm0.01$&$0.063\pm0.011$&$0.062\pm0.018$&$0.064\pm0.016$&$0.107\pm0.031$\\ \hline

 & \texttt{SyN} & \texttt{SPM5N8} & \texttt{SPM5N} & \texttt{SPM5U} & \texttt{SPM5D} & \texttt{LO} & \texttt{LP} & \texttt{LPC} \\ \hline
\rowcolor{green!30} \cellcolor{white!30}\texttt{LO}&$0.020\pm0.011$&$0.169\pm0.029$&$0.114\pm0.019$&$0.1\pm0.015$& $0.022\pm0.049$& \cellcolor{white!30}N/A&$0.02\pm0.004$&\cellcolor{green!30}$0.009\pm0.002$\\ \hline
\rowcolor{green!30} \cellcolor{white!30}\texttt{LP}&\cellcolor{blue!30}$0.001\pm0.011$&$0.149\pm0.028$&$0.095\pm0.017$&$0.076\pm0.013$&\cellcolor{blue!30}$0.003\pm0.048$&\cellcolor{red!30}$-0.02\pm0.004$& \cellcolor{white!30}N/A&\cellcolor{red!30}$-0.011\pm0.003$\\ \hline
\rowcolor{green!30} \cellcolor{white!30}\texttt{LPC}&$0.012\pm0.011$&$0.16\pm0.028$&$0.106\pm0.018$&$0.087\pm0.013$&$\cellcolor{blue!30}0.013\pm0.048$&\cellcolor{red!30}$0.009\pm0.002$&$0.011\pm0.003$& \cellcolor{white!30}N/A\\ \hline

\end{tabularx}
\end{center}
\vskip2ex
\end{subtable}

\begin{subtable}{1\linewidth}
\begin{center}
\begin{tabularx}{1\textwidth}{|*{9}{>{\hsize=0.11\linewidth}X|}}
\hline
\multicolumn{9}{|c|}{\textbf{Dataset}: \texttt{MGH10}} \\ \hline\hline
 & \texttt{FLIRT} & \texttt{AIR} & \texttt{ANIMAL} & \texttt{ART} & \texttt{Demons} & \texttt{FNIRT} & \texttt{Fluid} & \texttt{SLICE} \\ \hline
\rowcolor{green!30} \cellcolor{white!30}\texttt{LO}&$0.104\pm0.016$&$0.087\pm0.015$&$0.062\pm0.022$&$\cellcolor{blue!30}0.005\pm0.016$&$0.044\pm0.013$&$0.071\pm0.018$&$0.043\pm0.016$&$0.083\pm0.017$\\ \hline
\rowcolor{green!30} \cellcolor{white!30}\texttt{LP}&$0.091\pm0.016$&$0.073\pm0.016$&$0.049\pm0.023$&\cellcolor{red!30}$-0.008\pm0.017$&$0.03\pm0.013$&$0.058\pm0.018$&$0.03\pm0.015$&$0.07\pm0.017$\\ \hline
\rowcolor{green!30} \cellcolor{white!30}\texttt{LPC}&$0.098\pm0.016$&$0.081\pm0.015$&$0.057\pm0.022$&\cellcolor{blue!30}$0\pm0.016$&$0.038\pm0.013$&$0.065\pm0.018$&$0.037\pm0.016$&$0.077\pm0.017$\\ \hline

 & \texttt{SyN} & \texttt{SPM5N8} & \texttt{SPM5N} & \texttt{SPM5U} & \texttt{SPM5D} & \texttt{LO} & \texttt{LP} & \texttt{LPC} \\ \hline
\rowcolor{green!30} \cellcolor{white!30}\texttt{LO}&\cellcolor{blue!30}$-0.002\pm0.015$&$0.069\pm0.02$&$0.135\pm0.041$&$0.07\pm0.024$&$0.023\pm0.047$& \cellcolor{white!30}N/A&$0.013\pm0.004$&$0.006\pm0.002$\\ \hline
\rowcolor{green!30} \cellcolor{white!30}\texttt{LP}&\cellcolor{red!30}$-0.015\pm0.016$&$0.055\pm0.02$&$0.121\pm0.041$&$0.057\pm0.025$&$0.01\pm0.048$&\cellcolor{red!30}$\cellcolor{blue!30}-0.013\pm0.004$& \cellcolor{white!30}N/A&\cellcolor{red!30}$-0.008\pm0.003$\\ \hline
\rowcolor{green!30} \cellcolor{white!30}\texttt{LPC}&\cellcolor{red!30}$-0.008\pm0.015$&$0.063\pm0.02$&$0.129\pm0.041$&$0.065\pm0.024$&$0.018\pm0.047$&\cellcolor{red!30}$\cellcolor{blue!30}-0.006\pm0.002$&$0.008\pm0.003$& \cellcolor{white!30}N/A\\ \hline

\end{tabularx}
\end{center}
\vskip2ex
\end{subtable}

\vskip4ex
\caption{\label{table:ttest} Mean and standard deviation of the difference of target overlap score between LDDMM variants (LDDMM optimization (\texttt{LO}), the proposed prediction network (\texttt{LP}) and prediction+correction network (\texttt{LPC})) and all other methods for the \emph{image-to-image} experiments. The cell coloring indicates significant differences calculated from a pair-wise $t$-test: \textcolor{green}{green} indicates that the row-method is statistically significantly \emph{better} than the column-method; \textcolor{red}{red} indicates that the row-method is statistically significantly \emph{worse} than the column-method, and \textcolor{blue}{blue} indicates the difference is not statistically significant (best-viewed in color). \xyrevision{We use Bonferroni correction to safe-guard against spurious results due to multiple comparisons by dividing the significance level $\alpha$ by 204 (the total number of statistical tests).} \mnr{The significance level for rejection of the null-hypothesis is $\alpha=0.05/204$. Best-viewed in color.}}
\end{table*}

\vskip1ex

\begin{table*}[t!]
\centering
\scriptsize
\begin{subtable}{0.495\linewidth}
\begin{center}
\begin{tabular}{|r|c|c|c|c|c|}
\hline
\multicolumn{6}{|c|}{\textbf{Dataset}: \texttt{LPBA40}} \\ \hline
& \texttt{ART} & \texttt{SyN}  & \texttt{LO} &\texttt{LP}& \texttt{LPC}\\ \hline
\texttt{LO} & & & N/A & \cellcolor{green!30}$\checkmark$ & \cellcolor{green!30}$\checkmark$\\ \hline
\texttt{LP} & & & \cellcolor{green!30}$\checkmark$ & N/A & \cellcolor{green!30}$\checkmark$\\ \hline
\texttt{LPC} & & & \cellcolor{green!30}$\checkmark$ & \cellcolor{green!30}$\checkmark$  & N/A\\ \hline
\end{tabular}
\end{center}
\end{subtable}
\begin{subtable}{0.495\linewidth}
\begin{center}
\begin{tabular}{|r|c|c|c|c|c|}
\hline
\multicolumn{6}{|c|}{\textbf{Dataset}: \texttt{IBSR18}} \\ \hline
& \texttt{ART} & \texttt{SyN}  & \texttt{LO} &\texttt{LP}& \texttt{LPC}\\ \hline
\texttt{LO} & &\cellcolor{green!30}$\checkmark$ & N/A & & \cellcolor{green!30}$\checkmark$\\ \hline
\texttt{LP} &\cellcolor{green!30}$\checkmark$ & & & N/A & \\ \hline
\texttt{LPC} & &\cellcolor{green!30}$\checkmark$ & \cellcolor{green!30}$\checkmark$ & & N/A\\ \hline
\end{tabular}
\end{center}
\end{subtable}
\begin{subtable}{0.495\linewidth}
\begin{center}
\begin{tabular}{|r|c|c|c|c|c|}
\hline
\multicolumn{6}{|c|}{\textbf{Dataset}: \texttt{CUMC12}} \\ \hline
& \texttt{ART} & \texttt{SyN}  & \texttt{LO} &\texttt{LP}& \texttt{LPC}\\ \hline
\texttt{LO} & & & N/A & & \cellcolor{green!30}$\checkmark$\\ \hline
\texttt{LP} & & \cellcolor{green!30}$\checkmark$ & & N/A & \\ \hline
\texttt{LPC} & & & \cellcolor{green!30}$\checkmark$ &   & N/A\\ \hline
\end{tabular}
\end{center}
\end{subtable}
\begin{subtable}{0.495\linewidth}
\begin{center}
\begin{tabular}{|r|c|c|c|c|c|}
\hline
\multicolumn{6}{|c|}{\textbf{Dataset}: \texttt{MGH10}} \\ \hline
& \texttt{ART} & \texttt{SyN}  & \texttt{LO} &\texttt{LP}& \texttt{LPC}\\ \hline
\texttt{LO} & & \cellcolor{green!30}$\checkmark$ & N/A & &\cellcolor{green!30}$\checkmark$ \\ \hline
\texttt{LP} & & & & N/A & \cellcolor{green!30}$\checkmark$\\ \hline
\texttt{LPC} & \cellcolor{green!30}$\checkmark$ & &\cellcolor{green!30}$\checkmark$  &\cellcolor{green!30}$\checkmark$  & N/A\\ \hline
\end{tabular}
\end{center}
\end{subtable}
\vskip4ex
\caption{\label{table:TOST} Pairwise TOST, where we test the null-hypothesis that for the target overlap score for each row-method, $t_{\text{row}}$, and the target overlap score for each column-method, $t_{\text{column}}$, $\frac{t_{\text{row}}}{t_{\text{column}}} < 0.98$, or $\frac{t_{\text{row}}}{t_{\text{column}}} > 1.02$. Rejecting the null-hypothesis indicates that the row-method and column-method are statistically equivalent. Equivalence is marked as $\checkmark$s in the table. \xyrevision{We use Bonferroni correction to safe-guard against spurious results due to multiple comparisons by dividing the significance level $\alpha$ by 204 (the total number of statistical tests).} \mnr{The significance level for rejection of the null-hypothesis is $\alpha=0.05/204$.}}
\end{table*}

{To study the differences \rknew{among} registration algorithms {\it statistically}, we performed paired $t$-tests\footnote{\xyrevision{To safe-guard against overly optimistic results due to multiple comparisons, we used Bonferroni correction for all statistical tests in the paper (paired $t$-tests and paired TOST) by dividing the significance level $\alpha$ by the total number (204) of statistical tests we performed. This resulted in an effective significance level $\alpha=0.05/204\approx 0.00025$. The Bonferroni correction is likely overly strict for our experiments as the different registration results will be highly correlated, because they are based on the same input data.}} with respect to the target overlap scores between our LDDMM variants (\texttt{LO}, \texttt{LP}, \texttt{LPC}) and the methods in~\cite{Klein2009786}. Our null-hypothesis is that the methods show the same target overlap scores. We use a significance level of $\alpha=0.05/204$ for rejection of this null-hypothesis. We also computed the mean and the standard deviation of pair-wise differences between our LDDMM variants and these other methods. Table~\ref{table:ttest} shows the results. We observe that direct numerical optimization of the shooting LDDMM formulation via \texttt{PyCA} (\texttt{LO}) is a highly competitive registration method and shows better target overlap scores than most of the other registration algorithms for all four datasets (\texttt{LPBA40}, \texttt{IBSR18}, \texttt{CUMC12}, and \texttt{MGH10}). Notable exceptions are \texttt{ART} (on \texttt{LPBA40}), \texttt{SyN} (on \texttt{LBPA40}), and \texttt{SPM5D} (on \texttt{IBSR18}). However, performance decreases are generally very small: $-0.017$, $-0.013$, and $-0.009$ mean decrease in target overlap ratio for the three aforementioned exceptions, respectively. Specifically, a similar performance of \texttt{LO} to \texttt{SyN}, for example, is expected as \texttt{SyN} (as used in~\cite{Klein2009786}) is based on a {\it relaxation} formulation of LDDMM, whereas \texttt{LO} is based on the shooting formulation of LDDMM. Performance differences may be due to differences in the used regularizer and the image similarity measure. In particular, where \texttt{SyN} was used with Gaussian smoothing and cross-correlation, we used SSD as the image similarity measure and a regularizer involving up to second order spatial derivatives.

  \texttt{LO} is the algorithm that our predictive registration approaches (\texttt{LP} and \texttt{LPC}) are based on. Hence, \texttt{LP} and \texttt{LPC} are not expected to show improved performance with respect to \texttt{LO}. However, similar performance for \texttt{LP} and \texttt{LPC} would indicate high quality predictions. Indeed, Table~\ref{table:ttest} shows that our prediction+correction approach (\texttt{LPC}) performs similar (with respect to the other registration methods) to \texttt{LO}. A slight performance drop with respect to $\texttt{LO}$ can be observed for \texttt{LPC} and a slightly bigger performance drop for \texttt{LP}, which only uses \rknew{the} prediction model, but no correction model. 

To assess statistical equivalence of the top performing registration algorithms we performed paired two one-sided tests (paired TOST)~\cite{wellek2010testing} with a relative threshold difference of 2\%. In other words, our null-hypothesis is that methods show a relative difference of larger than 2\%. Rejection of this null-hypothesis at a significance level of $\alpha=0.05/204$ then \rknew{indicates
evidence} for statistical equivalence. Table~\ref{table:TOST} shows the paired TOST results. For a relative threshold difference of 2\% \texttt{LPC} can be considered statistically equivalent to \texttt{LO} for all four datasets and to many of the other top methods (e.g., \texttt{LPC} \emph{vs.} SyN on \texttt{MGH10} and \texttt{IBSR18}).

Overall, these statistical tests confirm that our prediction models, in particular \texttt{LPC}, are highly competitive registration algorithms. Computational cost, however, is very small. This is discussed in detail in Sec.~\ref{sec:runtime}.

\xyrevision{
\subsubsection{Choosing the correct ``correction step''}
\label{subsubsec:correct}
As shown in Sec.~\ref{subsubsec:overlap}, \texttt{LPP}, \texttt{LPC2} and \texttt{LPC3} all result in label overlapping scores which are similar to the label overlapping scores obtained via \texttt{LPC}. This raises the question which method should be preferred for the correction step. Note that among these methods, only \texttt{LPC} is specifically trained to match the LDDMM optimization results and in particular to predict {\it corrections} to the initial momentum obtained by the prediction model (\texttt{LP}) in the tangent space of the moving image. In contrast, \texttt{LPP}, \texttt{LPC2} and \texttt{LPC3} lack this theoretical motivation. Hence, it is unclear for these methods what the overall optimization goal is. To show what this means in practice, we computed the determinant of the Jacobian of the deformation maps ($\Phi^{-1}$) for all voxels for all four registration cases of~\cite{Klein2009786} inside the brain mask and calculated the histogram of the computed values. Our goal is to check the similarity (in distribution) between deformations generated by the prediction models (\texttt{LP}, \texttt{LPC}, \texttt{LPP}, \texttt{LPC2}, \texttt{LPC3}) in comparison to the results obtained via numerical LDDMM optimization (\texttt{LO}). 

As an example, Fig.~\ref{fig:distribution_jac} shows the result for the \texttt{LPBA40} dataset. The other three datasets show similar results. Fig.~\ref{fig:distribution_jac}(left) shows the histogram of the logarithmically transformed determinant of the Jacobian ($\text{log}_{10}\text{det}J$) for all the methods. A value of 0 on the x-axis indicates no deformation or a volume preserving deformation, $> 0$ indicates volumetric shrinkage and $< 0$ indicates volumetric expansion. We can see that \texttt{LPC} is closest to \texttt{LO}. \texttt{LP} generates smoother deformations compared with \texttt{LO}, which is sensible as one-step predictions will likely not be highly accurate and, in particular, may result in predicted momenta which are slightly smoother than the ones obtained by numerical optimization. Hence, in effect, the predictions may result in a more strongly spatially regularized deformation. \texttt{LPP}, \texttt{LPC2} and \texttt{LPC3} generate more drastic deformations (i.e., more spread out histograms indicating areas of stronger expansions and contractions). Fig.~\ref{fig:distribution_jac}(right) shows this effect more clearly; it shows the differences between the histogram of the prediction models and the registration result obtained by numerical optimization (\texttt{LO}). Hence, a method which is similar to \texttt{LO} in distribution will show a curve close to $y=0$.

This \rknew{assessment} also demonstrates that the correction network (of \texttt{LPC}) is different from the prediction network (\texttt{LP}): the correction network is trained specifically to correct \textsl{minor errors} in the predicted momenta of the prediction network with respect to the desired momenta obtained by numerical optimization (\texttt{LO}), while the prediction network is not. Thus, \texttt{LPC} is the only model among the prediction models (apart from \texttt{LP}) that has the explicit goal of predicting the behavior of the LDDMM optimization result (\texttt{LO}). When we use the prediction network in the correction step, the high label overlapping scores are due to more drastic deformations compared with \texttt{LP}, but there is no clear theoretical justification of \texttt{LPP}. In fact, it is more reminiscent of a greedy solution strategy, albeit still results in geodesic paths as the predicted momenta are added in the tangent space of the undeformed moving image. Similar arguments hold for \texttt{LPC2} and \texttt{LPC3}: using the correction network multiple times (iteratively) in the correction step also results in increasingly drastic deformations, as illustrated by the curves for \texttt{LPC}, \texttt{LPC2} and \texttt{LPC3} in Fig.~\ref{fig:distribution_jac}. Compared to the label overlapping accuracy boost from \texttt{LP} to \texttt{LPC}, \texttt{LPC2} and \texttt{LPC3} do not greatly improve the registration accuracy, and may even generate worse results (e.g., \texttt{LPC3} on \texttt{LPBA40}). Furthermore, the additional computation cost for more iterations of the correction network + LDDMM shooting makes \texttt{LPC2} and \texttt{LPC3} less favorable, in comparison to \texttt{LPC}.

\begin{figure*}[htbp]
    \begin{center}
        \includegraphics[width=\textwidth]{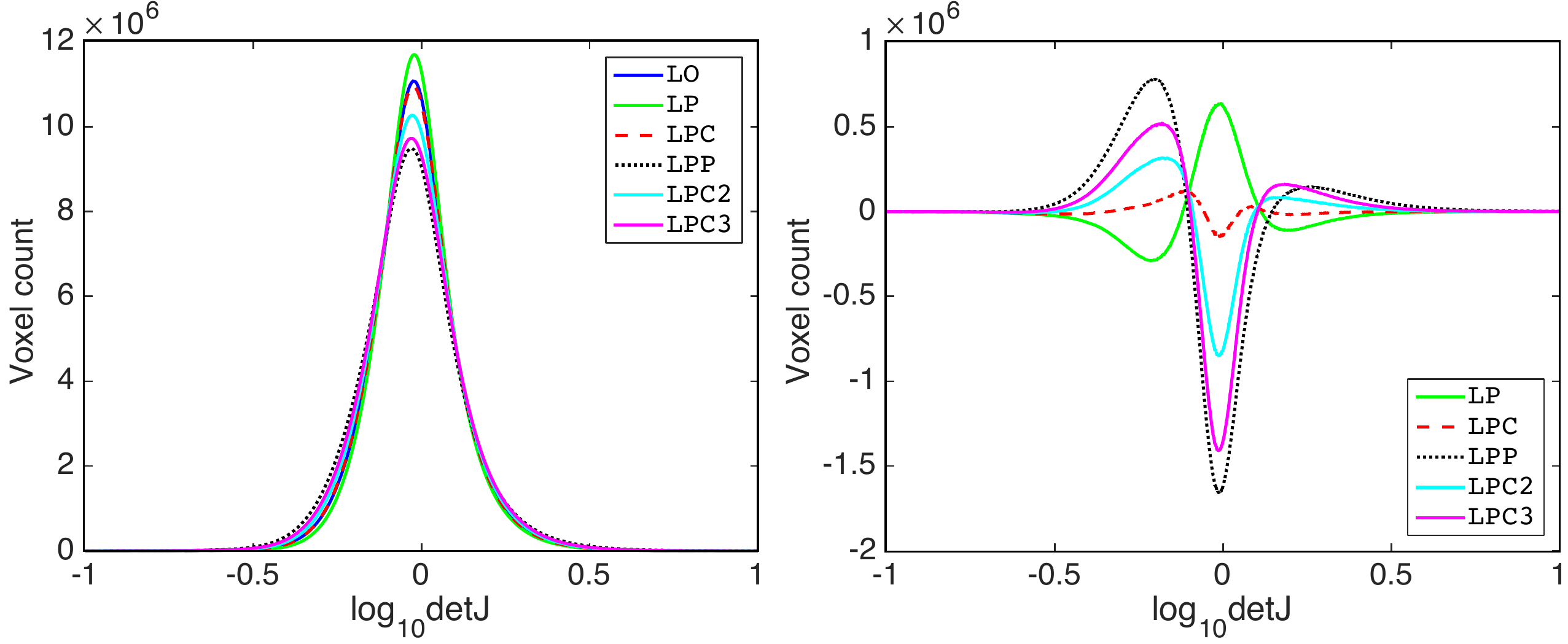}
    \end{center}
    \caption{\label{fig:distribution_jac}\xyrevision{Distribution of the determinant of Jacobian of the deformations for \texttt{LPBA40} dataset registrations. \emph{Left}: histograms of the log-transformed determinant of Jacobian for the deformation maps ($\text{log}_{10}\text{det}J$) for all registration cases. \emph{Right}: difference of the histograms of $\text{log}_{10}\text{det}J$ between prediction models (\texttt{LP}, \texttt{LPC}, \texttt{LPP}, \texttt{LPC2}, \texttt{LPC3}) and \texttt{LO}. For the right figure, the closer a curve is to $y=0$, the more similar the corresponding method is to \texttt{LO}. A \rknew{value of 0 on the $x$-axis} indicates no deformation, or a volume-preserving deformation, $> 0$ indicates shrinkage and $< 0$ indicates expansion. Best-viewed in color.}}
\end{figure*}

\subsubsection{Predicting various ranges of deformations}
\label{subsubsec:deform}

Table~\ref{table:def_range} shows the range of deformations and associated percentiles for the deformation fields generated by LDDMM optimization for the four \textsl{image-to-image} test datasets. \mnr{All computations were restricted to locations inside the brain mask.} Table~\ref{table:def_range} also shows the means and standard deviations of the differences of deformations between the results for the prediction models and the results obtained by numerical optimization (\texttt{LO}). As shown in the table, the largest deformations that LDDMM optimization generates are 23.393~mm for \texttt{LPBA40}, 36.263~mm for \texttt{IBSR18}, 18.753~mm for \texttt{CUMC12} and 18.727~mm for \texttt{MGH10}.

Among the prediction models, \texttt{LPC} improves the prediction accuracy compared with \texttt{LP}, and generally achieves the highest deformation prediction accuracy for up to 80\% of the voxels. It is also on-par with other prediction models for up to 99\% of the voxels, where the largest deformations are in the range between 7.317~mm-9.026~mm for the four datasets. For very large deformations that occur for 1\% of the total voxels, \texttt{LPC} does not drastically reduce the deformation error. This is due to the following three reasons: \emph{First}, the input patch size of the deep learning framework is $15\times15\times15$, which means that the receptive field for the network input is limited to $15\times15\times15\text{mm}^3$. This constrains the network's ability to predict very large deformations, and can potentially be solved by implementing a multi-scale input network for prediction. \emph{Second}, the deformations in the OASIS training images \mnr{have a median of 2.609~mm, which is similar to the median observed in the four testing datasets. However, only 0.2\% of the voxels in the OASIS training dataset have deformations larger than 10~mm. Such a small number of training patches containing very large deformations makes it difficult to train the network to accurately predict these very large deformations in the test data. \mnr{If capturing these very large deformations is desired, a possible solution could be to provide a larger number of training examples for large deformations or to weight samples based on their importance.} \mnr{\emph{Third}, outliers in the dataset whose appearances are very different from the other images in the dataset can cause very large deformations. For example, in the \texttt{IBSR18} dataset, only three distinct images are needed as moving or target images to cover the 49 registration cases that generate deformations larger than 20~mm. These large deformations created by numerical LDDMM optimization are not always desirable; and consequentially registration errors of the prediction models with respect to the numerical optimization result are in fact sometimes preferred. As a case in point, Fig.~\ref{fig:fail} shows a registration failure case from the \texttt{IBSR18} dataset for LDDMM optimization and the corresponding prediction result. In this example, the brain extraction did not extract consistent anatomy for the moving image and the target image. Specifically, only inconsistent parts of the cerebellum remain between the moving and the target images. As optimization-based LDDMM does not know about this inconsistency, it attempts to match the images as well as possible and thereby creates a very extreme deformation. Our prediction result, however, still generates reasonable deformations (where plausibility is based on the deformations that were observed during training) while matching the brain structures as much as possible. This can be regarded as an \textsl{advantage} of our network, where the conservative nature of patch-wise momentum prediction is more likely to generate reasonable deformations.}
}

\begin{figure}[htbp]
\centering
\includegraphics[width=0.48\textwidth]{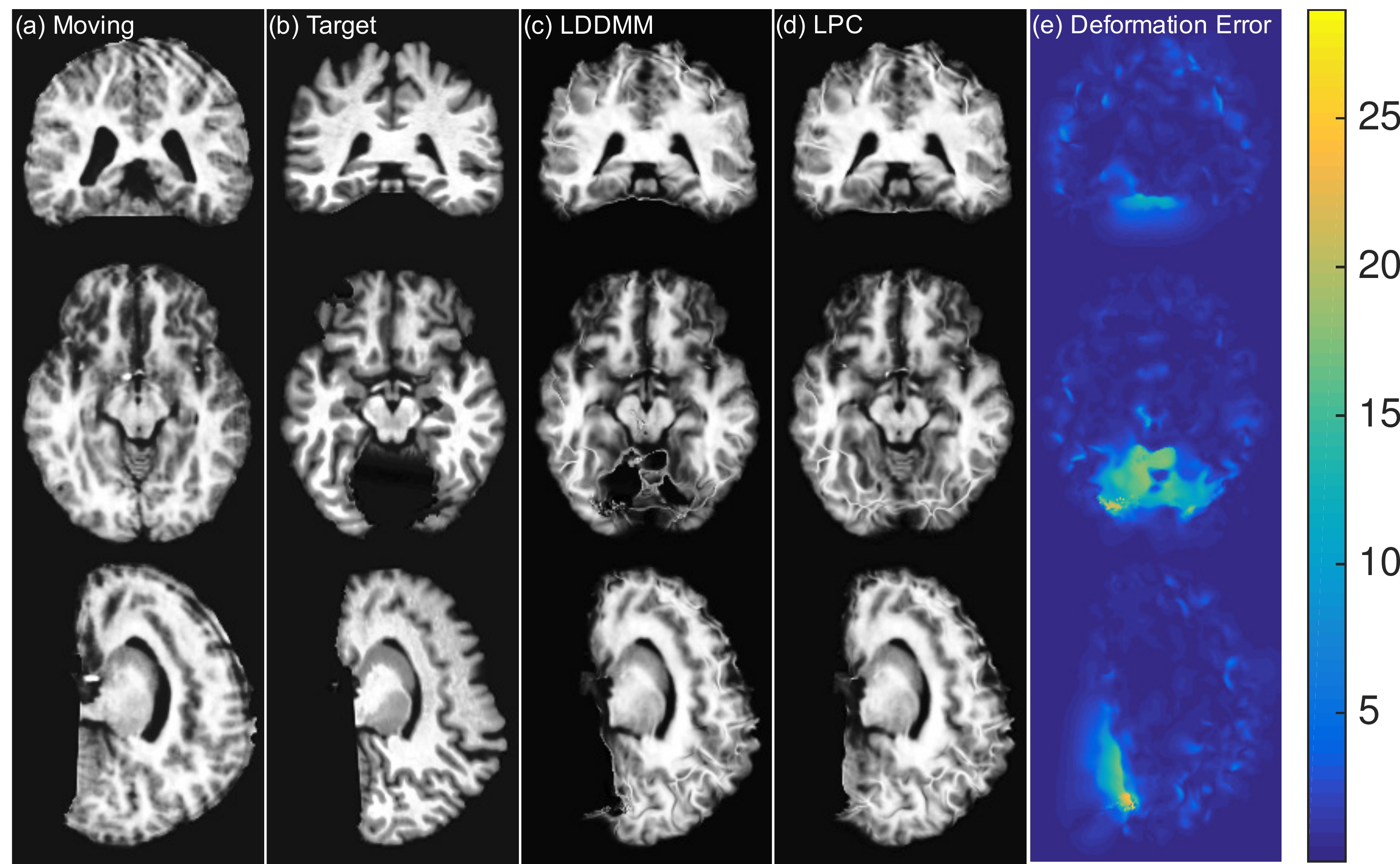}
\vskip 2ex
\caption{\label{fig:fail}\xyrevision{Failure case for \texttt{IBSR18} dataset where LDDMM optimization generated very extreme deformations. From \emph{left} to \emph{right}: (a): moving image; (b): target image; (c): LDDMM optimization result; (d): prediction+correction result (\texttt{LPC}); (e): heatmap showing the differences between the optimization deformation and predicted deformation in millimeters. \mnr{Most registration errors occur in the area of the cerebellum, which has been inconsistently preserved in the moving and the target images during brain extraction. Hence, not all the retained brain regions in the moving image have correspondences in the target image. Best-viewed in color.}}}
\end{figure}

\begin{table*}[htbp]
\fontsize{6}{6}\selectfont
\begin{subtable}{1\linewidth}
\setlength\tabcolsep{1pt}
\begin{center}
\begin{tabularx}{1\textwidth}{|*{9}{>{\hsize=0.11\linewidth}X|}}
\hline
\multicolumn{9}{|c|}{\scriptsize \textbf{Dataset}: \texttt{LPBA40}} \\ \hline\hline
Voxel\%& 0\%-10\% & 10\%-20\% & 20\%-30\% & 30\%-40\% & 40\%-50\% & 50\%-60\% & 60\%-70\% & 70\%-80\% \\ \hline
\# of Test Cases & 1560(100\%) & 1560(100\%) & 1560(100\%) & 1560(100\%) & 1560(100\%) & 1560(100\%) & 1560(100\%) & 1560(100\%)\\\hline
$I_{\text{main}}$ & 40 & 40 & 40 & 40 & 40 & 40 & 40 & 40 \\ \hline
Deform (mm) & 0.001-0.999&0.999-1.400&1.400-1.751&1.751-2.093&2.093-2.450&2.450-2.840&2.840-3.295&3.295-3.873\\\hline
\texttt{LP} & $0.478\pm0.323$ &$0.564\pm0.361$ &$0.623\pm0.393$ &$0.675\pm0.425$ &$0.728\pm0.461$ &$0.786\pm0.503$ &$0.853\pm0.556$ &$0.941\pm0.630$ \\\hline
\texttt{LPC} & $\cellcolor{green!30} 0.415\pm0.287$ & \cellcolor{green!30}$0.469\pm0.327$ &\cellcolor{green!30}$0.509\pm0.359$ &\cellcolor{green!30}$0.546\pm0.390$ &\cellcolor{green!30}$0.584\pm0.424$ &\cellcolor{green!30}$0.626\pm0.464$ &\cellcolor{green!30}$0.676\pm0.514$ &\cellcolor{green!30}$0.742\pm0.583$ \\ \hline
\texttt{LPP} & $0.543\pm0.554$ &$0.605\pm0.613$ &$0.650\pm0.684$ &$0.691\pm0.823$ &$0.733\pm0.783$ &$0.778\pm0.852$ &$0.830\pm0.922$ &$0.898\pm1.055$ \\\hline
\texttt{LPC2} & $0.510\pm0.339$ &$0.551\pm0.378$ &$0.582\pm0.409$ &$0.611\pm0.439$ &$0.642\pm0.471$ &$0.676\pm0.508$ &$0.716\pm0.553$ &$0.770\pm0.617$ \\\hline
\texttt{LPC3} & $0.637\pm0.410$ &$0.674\pm0.451$ &$0.703\pm0.486$ &$0.732\pm0.516$ &$0.762\pm0.550$ &$0.795\pm0.588$ &$0.834\pm0.638$ &$0.886\pm0.713$ \\\hline\hline

Voxel\%& 80\%-90\% & 90\%-99\% & 99\%-100\% & 99.9\%-100\% & 99.99\%-100\% & 99.999\%-100\% & 99.9999\%-100\% & 99.99999\%-100\% \\ \hline
\# of Test Cases & 1560(100\%) & 1560(100\%) & 1560(100\%) & 1474(94.5\%) & 417(26.7\%) & 72(4.6\%) & 30(1.9\%) & 8(0.5\%)\\\hline
$I_{\text{main}}$ & 40 & 40 & 40 & 38 & 31 & 8 & 2 & 1 \\ \hline
Deform (mm) &3.873-4.757&4.757-7.317&7.317-23.393&9.866-23.393&12.435-23.393&14.734-23.393&16.835-23.393&19.090-23.393\\\hline
\texttt{LP} & $1.079\pm0.752$ &$1.418\pm1.056$ &$2.579\pm1.869$ &$4.395\pm2.667$ &$6.863\pm3.657$ &$9.220\pm4.829$ &$11.568\pm6.340$ &$14.475\pm7.145$ \\\hline
\texttt{LPC} & \cellcolor{green!30}$0.847\pm0.696$ &$1.101\pm0.976$ &$1.961\pm1.761$ &$3.431\pm2.656$ &$5.855\pm3.711$ &$8.408\pm4.780$ &$\cellcolor{green!30}10.484\pm6.148$ &$14.041\pm6.879$ \\\hline
\texttt{LPP} & $1.001\pm1.396$ &$1.229\pm1.558$ &\cellcolor{green!30}$1.760\pm2.965$ &\cellcolor{green!30}$2.514\pm5.845$ &\cellcolor{green!30}$4.127\pm7.988$ &\cellcolor{green!30}$7.097\pm3.566$ &$10.509\pm5.093$ &\cellcolor{green!30}$11.436\pm5.688$ \\\hline
\texttt{LPC2} & $0.856\pm0.721$ &\cellcolor{green!30}$1.064\pm0.979$ &$1.765\pm1.726$ &$3.012\pm2.644$ &$5.351\pm3.698$ &$8.362\pm4.765$ &$11.310\pm6.117$ &$15.957\pm6.458$ \\\hline
\texttt{LPC3} & $0.968\pm0.835$ &$1.166\pm1.176$ &$1.829\pm2.880$ &$3.019\pm4.688$ &$5.680\pm10.450$ &$11.434\pm23.727$ &$18.939\pm33.953$ &$24.152\pm38.466$ \\ \hline
\end{tabularx}
\end{center}
\vskip1ex
\end{subtable}

\begin{subtable}{1\linewidth}
\setlength\tabcolsep{1pt}
\begin{center}
\begin{tabularx}{1\textwidth}{|*{9}{>{\hsize=0.11\linewidth}X|}}
\hline
\multicolumn{9}{|c|}{\scriptsize \textbf{Dataset}: \texttt{IBSR18}} \\ \hline\hline
Voxel\%& 0\%-10\% & 10\%-20\% & 20\%-30\% & 30\%-40\% & 40\%-50\% & 50\%-60\% & 60\%-70\% & 70\%-80\% \\ \hline
\# of Test Cases & 306(100\%) & 306(100\%) & 306(100\%) & 306(100\%) & 306(100\%) & 306(100\%) & 306(100\%) & 306(100\%)\\\hline
$I_{\text{main}}$ & 18 & 18 & 18 & 18 & 18 & 18 & 18 & 18 \\ \hline
Deform (mm) & 0.003-1.221&1.221-1.691&1.691-2.101&2.101-2.501&2.501-2.915&2.915-3.370&3.370-3.901&3.901-4.580\\\hline
\texttt{LP} & $0.573\pm0.401$ &$0.670\pm0.448$ &$0.742\pm0.489$ &$0.810\pm0.532$ &$0.879\pm0.580$ &$0.957\pm0.637$ &$1.053\pm0.711$ &$1.182\pm0.815$ \\\hline
\texttt{LPC} & \cellcolor{green!30}$0.450\pm0.343$ &\cellcolor{green!30}$0.521\pm0.393$ &\cellcolor{green!30}$0.577\pm0.435$ &\cellcolor{green!30}$0.631\pm0.479$ &\cellcolor{green!30}$0.687\pm0.526$ &\cellcolor{green!30}$0.751\pm0.582$ &$0.829\pm0.655$ &$0.936\pm0.756$ \\\hline
\texttt{LPP} & $0.624\pm0.553$ &$0.698\pm0.623$ &$0.755\pm0.678$ &$0.809\pm0.729$ &$0.863\pm0.782$ &$0.923\pm0.843$ &$0.996\pm0.918$ &$1.094\pm1.022$ \\\hline
\texttt{LPC2} & $0.505\pm0.383$ &$0.566\pm0.437$ &$0.614\pm0.480$ &$0.660\pm0.522$ &$0.707\pm0.568$ &$0.761\pm0.622$ &$\cellcolor{green!30}0.827\pm0.691$ &\cellcolor{green!30}$0.917\pm0.788$ \\\hline
\texttt{LPC3} & $0.606\pm0.446$ &$0.666\pm0.504$ &$0.713\pm0.549$ &$0.757\pm0.594$ &$0.803\pm0.640$ &$0.854\pm0.694$ &$0.915\pm0.763$ &$1.001\pm0.860$ \\\hline\hline

Voxel\%& 80\%-90\% & 90\%-99\% & 99\%-100\% & 99.9\%-100\% & 99.99\%-100\% & 99.999\%-100\% & 99.9999\%-100\% & 99.99999\%-100\% \\ \hline
\# of Test Cases & 306(100\%) & 306(100\%) & 306(100\%) & 125(40.8\%) & 46(15.0\%) & 12(3.9\%) & 3(1.0\%) & 3(1.0\%)\\\hline
$I_{\text{main}}$ & 18 & 18 & 18 & 10 & 3 & 2 & 1 & 1 \\ \hline
Deform (mm) & 4.580-5.629&5.629-9.026&9.026-36.263&14.306-36.263&19.527-36.263&23.725-36.263&27.533-36.263&29.154-36.263\\\hline
\texttt{LP} & $1.397\pm0.988$ &$2.007\pm1.451$ &$5.343\pm3.868$ &$13.103\pm4.171$ &$20.302\pm2.287$ &$24.666\pm1.688$ &$28.081\pm1.190$ &$30.436\pm1.807$ \\\hline
\texttt{LPC} & $1.114\pm0.924$ &$1.609\pm1.367$ &$4.485\pm3.862$ &$11.928\pm4.800$ &$19.939\pm2.549$ &$24.457\pm1.877$ &$27.983\pm1.218$ &$30.000\pm1.796$ \\\hline
\texttt{LPP} & $1.249\pm1.184$ &$1.621\pm1.572$ &\cellcolor{green!30}$2.894\pm2.798$ &\cellcolor{green!30}$5.054\pm4.387$ &\cellcolor{green!30}$9.631\pm6.406$ &\cellcolor{green!30}$13.541\pm7.744$ &\cellcolor{green!30}$17.600\pm7.317$ &\cellcolor{green!30}$15.553\pm6.339$ \\\hline
\texttt{LPC2} & \cellcolor{green!30}$1.068\pm0.951$ &\cellcolor{green!30}$1.488\pm1.375$ &$3.917\pm3.730$ &$10.437\pm5.395$ &$19.345\pm3.061$ &$24.154\pm2.173$ &$27.834\pm1.403$ &$29.654\pm1.909$ \\\hline
\texttt{LPC3} & $1.141\pm1.022$ &$1.521\pm1.432$ &$3.596\pm3.555$ &$8.962\pm5.690$ &$18.252\pm4.077$ &$23.673\pm2.583$ &$27.681\pm1.641$ &$29.460\pm2.227$ \\\hline

\end{tabularx}
\end{center}
\vskip1ex
\end{subtable}

\begin{subtable}{1\linewidth}
\setlength\tabcolsep{1pt}
\begin{center}
\begin{tabularx}{1\textwidth}{|*{9}{>{\hsize=0.11\linewidth}X|}}
\hline
\multicolumn{9}{|c|}{\scriptsize \textbf{Dataset}: \texttt{CUMC12}} \\ \hline\hline
Voxel\%& 0\%-10\% & 10\%-20\% & 20\%-30\% & 30\%-40\% & 40\%-50\% & 50\%-60\% & 60\%-70\% & 70\%-80\% \\ \hline
\# of Test Cases & 132(100\%) & 132(100\%) & 132(100\%) & 132(100\%) & 132(100\%) & 132(100\%) & 132(100\%) & 132(100\%)\\\hline
$I_{\text{main}}$ & 12 & 12 & 12 & 12 & 12 & 12 & 12 & 12 \\ \hline
Deform (mm) & 0.004-1.169&1.169-1.602&1.602-1.977&1.977-2.341&2.341-2.717&2.717-3.126&3.126-3.597&3.597-4.189\\\hline
\texttt{LP} & $0.617\pm0.433$ &$0.709\pm0.480$ &$0.784\pm0.524$ &$0.856\pm0.570$ &$0.929\pm0.621$ &$1.010\pm0.680$ &$1.109\pm0.758$ &$1.239\pm0.868$ \\\hline
\texttt{LPC} & \cellcolor{green!30}$0.525\pm0.391$ &\cellcolor{green!30}$0.587\pm0.441$ &\cellcolor{green!30}$0.640\pm0.486$ &\cellcolor{green!30}$0.694\pm0.534$ &\cellcolor{green!30}$0.750\pm0.586$ &\cellcolor{green!30}$0.813\pm0.646$ &\cellcolor{green!30}$0.890\pm0.724$ &\cellcolor{green!30}$0.995\pm0.834$ \\\hline
\texttt{LPP} & $0.653\pm0.538$ &$0.717\pm0.607$ &$0.772\pm0.667$ &$0.829\pm0.727$ &$0.888\pm0.789$ &$0.953\pm0.859$ &$1.032\pm0.948$ &$1.138\pm1.068$ \\\hline
\texttt{LPC2} & $0.605\pm0.444$ &$0.653\pm0.496$ &$0.696\pm0.543$ &$0.739\pm0.591$ &$0.787\pm0.645$ &$0.839\pm0.704$ &$0.905\pm0.782$ &$0.996\pm0.891$ \\\hline
\texttt{LPC3} & $0.730\pm0.519$ &$0.775\pm0.575$ &$0.815\pm0.625$ &$0.857\pm0.675$ &$0.903\pm0.732$ &$0.954\pm0.793$ &$1.019\pm0.872$ &$1.107\pm0.984$ \\\hline\hline

Voxel\%& 80\%-90\% & 90\%-99\% & 99\%-100\% & 99.9\%-100\% & 99.99\%-100\% & 99.999\%-100\% & 99.9999\%-100\% & 99.99999\%-100\% \\ \hline
\# of Test Cases & 132(100\%) & 132(100\%) & 132(100\%) & 132(100\%) & 75(56.8\%) & 18(13.6\%) & 1(0.8\%) & 1(0.8\%)\\\hline
$I_{\text{main}}$ & 12 & 12 & 12 & 12 & 9 & 6 & 1 & 1 \\ \hline
Deform (mm) & 4.189-5.070&5.070-7.443&7.443-18.753&9.581-18.753&12.115-18.753&14.383-18.753&16.651-18.753&18.297-18.753\\\hline
\texttt{LP} & $1.448\pm1.050$ &$1.955\pm1.490$ &$3.340\pm2.412$ &$4.882\pm3.106$ &$7.281\pm3.378$ &$9.978\pm3.211$ &$14.113\pm1.211$ &$15.868\pm0.315$ \\\hline
\texttt{LPC} & $1.163\pm1.017$ &$1.571\pm1.451$ &$2.676\pm2.386$ &$3.930\pm3.125$ &$5.976\pm3.646$ &$8.527\pm3.889$ &$13.009\pm1.584$ &$14.908\pm0.437$ \\\hline
\texttt{LPP} & $1.305\pm1.258$ &$1.683\pm1.686$ &$2.484\pm2.503$ &\cellcolor{green!30}$3.047\pm3.048$ &\cellcolor{green!30}$3.511\pm3.287$ &\cellcolor{green!30}$2.999\pm3.150$ &\cellcolor{green!30}$1.196\pm0.327$ &\cellcolor{green!30}$1.277\pm0.178$ \\\hline
\texttt{LPC2} & \cellcolor{green!30}$1.142\pm1.072$ &\cellcolor{green!30}$1.494\pm1.494$ &$2.405\pm2.388$ &$3.392\pm3.055$ &$5.003\pm3.702$ &$7.176\pm4.218$ &$11.155\pm2.470$ &$13.456\pm0.812$ \\\hline
\texttt{LPC3} & $1.248\pm1.165$ &$1.581\pm1.581$ &\cellcolor{green!30}$2.383\pm2.434$ &$3.165\pm3.015$ &$4.422\pm3.629$ &$6.283\pm4.370$ &$9.607\pm3.120$ &$11.757\pm1.492$ \\\hline
\end{tabularx}
\end{center}
\vskip1ex
\end{subtable}

\begin{subtable}{1\linewidth}
\setlength\tabcolsep{1pt}
\begin{center}
\begin{tabularx}{1\textwidth}{|*{9}{>{\hsize=0.11\linewidth}X|}}
\hline
\multicolumn{9}{|c|}{\scriptsize \textbf{Dataset}: \texttt{MGH10}} \\ \hline\hline
Voxel\%& 0\%-10\% & 10\%-20\% & 20\%-30\% & 30\%-40\% & 40\%-50\% & 50\%-60\% & 60\%-70\% & 70\%-80\% \\ \hline
\# of Test Cases & 90(100\%) & 90(100\%) & 90(100\%) & 90(100\%) & 90(100\%) & 90(100\%) & 90(100\%) & 90(100\%)\\\hline
$I_{\text{main}}$ & 10 & 10 & 10 & 10 & 10 & 10 & 10 & 10 \\ \hline
Deform (mm) & 0.003-1.122&1.122-1.553&1.553-1.929&1.929-2.294&2.294-2.674&2.674-3.089&3.089-3.567&3.567-4.163\\\hline
\texttt{LP} & $0.578\pm0.422$ &$0.680\pm0.471$ &$0.757\pm0.514$ &$0.829\pm0.559$ &$0.904\pm0.610$ &$0.986\pm0.671$ &$1.082\pm0.748$ &$1.207\pm0.858$ \\\hline
\texttt{LPC} & \cellcolor{green!30}$0.486\pm0.382$ &\cellcolor{green!30}$0.558\pm0.436$ &\cellcolor{green!30}$0.615\pm0.481$ &\cellcolor{green!30}$0.669\pm0.528$ &\cellcolor{green!30}$0.726\pm0.580$ &\cellcolor{green!30}$0.790\pm0.640$ &\cellcolor{green!30}$0.865\pm0.715$ &\cellcolor{green!30}$0.963\pm0.820$ \\\hline
\texttt{LPP} & $0.624\pm0.556$ &$0.701\pm0.624$ &$0.761\pm0.680$ &$0.819\pm0.737$ &$0.881\pm0.796$ &$0.948\pm0.865$ &$1.026\pm0.948$ &$1.127\pm1.061$ \\\hline
\texttt{LPC2} & $0.562\pm0.436$ &$0.623\pm0.493$ &$0.670\pm0.541$ &$0.716\pm0.589$ &$0.766\pm0.643$ &$0.821\pm0.703$ &$0.886\pm0.778$ &$0.971\pm0.881$ \\\hline
\texttt{LPC3} & $0.684\pm0.512$ &$0.745\pm0.574$ &$0.792\pm0.625$ &$0.838\pm0.677$ &$0.887\pm0.734$ &$0.942\pm0.797$ &$1.007\pm0.874$ &$1.091\pm0.979$ \\\hline \hline

Voxel\%& 80\%-90\% & 90\%-99\% & 99\%-100\% & 99.9\%-100\% & 99.99\%-100\% & 99.999\%-100\% & 99.9999\%-100\% & 99.99999\%-100\% \\ \hline
\# of Test Cases & 90(100\%) & 90(100\%) & 90(100\%) & 89(98.9\%) & 37(41.1\%) & 7(7.8\%) & 3(3.3\%) & 2(2.2\%)\\\hline
$I_{\text{main}}$ & 10 & 10 & 10 & 9 & 6 & 3 & 1 & 1 \\ \hline
Deform (mm) & 4.163-5.047&5.047-7.462&7.462-18.727&9.833-18.727&12.607-18.727&15.564-18.727&17.684-18.727&18.534-18.727\\\hline
\texttt{LP} & $1.408\pm1.041$ &$1.924\pm1.499$ &$3.348\pm2.341$ &$4.873\pm2.704$ &$7.299\pm3.256$ &$10.503\pm4.049$ &$11.764\pm3.005$ &$13.041\pm1.691$ \\\hline
\texttt{LPC} & $1.120\pm0.998$ &$1.526\pm1.444$ &$2.627\pm2.275$ &$3.674\pm2.621$ &$5.283\pm3.272$ &$8.492\pm4.088$ &$9.336\pm3.738$ &$10.499\pm2.692$ \\\hline
\texttt{LPP} & $1.289\pm1.250$ &$1.676\pm1.705$ &$2.430\pm2.415$ &\cellcolor{green!30}$2.707\pm2.403$ &\cellcolor{green!30}$2.998\pm2.214$ &\cellcolor{green!30}$2.695\pm1.419$ &\cellcolor{green!30}$2.377\pm1.063$ &\cellcolor{green!30}$2.226\pm0.365$ \\\hline
\texttt{LPC2} & \cellcolor{green!30}$1.109\pm1.055$ &\cellcolor{green!30}$1.455\pm1.482$ &$2.361\pm2.270$ &$3.138\pm2.585$ &$4.267\pm3.133$ &$6.603\pm3.760$ &$7.444\pm3.657$ &$8.415\pm3.053$ \\\hline
\texttt{LPC3} & $1.225\pm1.153$ &$1.551\pm1.570$ &\cellcolor{green!30}$2.358\pm2.306$ &$3.023\pm2.588$ &$3.755\pm2.957$ &$5.214\pm3.486$ &$6.240\pm3.357$ &$7.569\pm3.283$ \\\hline

\end{tabularx}
\end{center}
\end{subtable}
\vskip 2ex
\caption{\label{table:def_range}\xyrevision{Deformation ranges and mean+standard \mnr{deviation} of the deformation errors between the prediction models (\texttt{LP, LPC, LPP, LPC2, LPC3}) and the optimization model (\texttt{LO}) for the \textsl{image-to-image} registration case. \mnr{All measures are evaluated within the brain mask only.} \textbf{All deformation values and deformation errors are evaluated in millimeters (mm)}. Voxel\%: percentile range of voxels that fall in a particular deformation range \mnr{based on the optimization model (\texttt{LO}). \# of Test cases: The number of registration cases that contain voxels within a given percentile range.}  \mnr{$I_{\text{main}}$: Minimum number of distinct images required to cover all registration test cases in a particular deformation range either as the moving or the target image. This measure is meant to quantify the influence of {\it few} images on very large deformations. For example, for the four registration cases A-B, B-C, B-D and E-A,  $I_{\text{main}}=2$ as it is sufficient to select images A and B to cover all four registrations. The results show that a comparatively small subset of images is responsible for most of the very large deformations. Of course, all images of a particular dataset are involved in the small deformation ranges. Deform: range of deformations within a given percentile range.} The cells with the lowest mean deformation errors for every deformation range are highlighted. Best-viewed in color.}}
\end{table*}

}
\begin{figure*}[t!]
    \begin{center}
    \begin{subfigure}[t]{0.5\textwidth}
        \centering
        \includegraphics[width=3.5in]{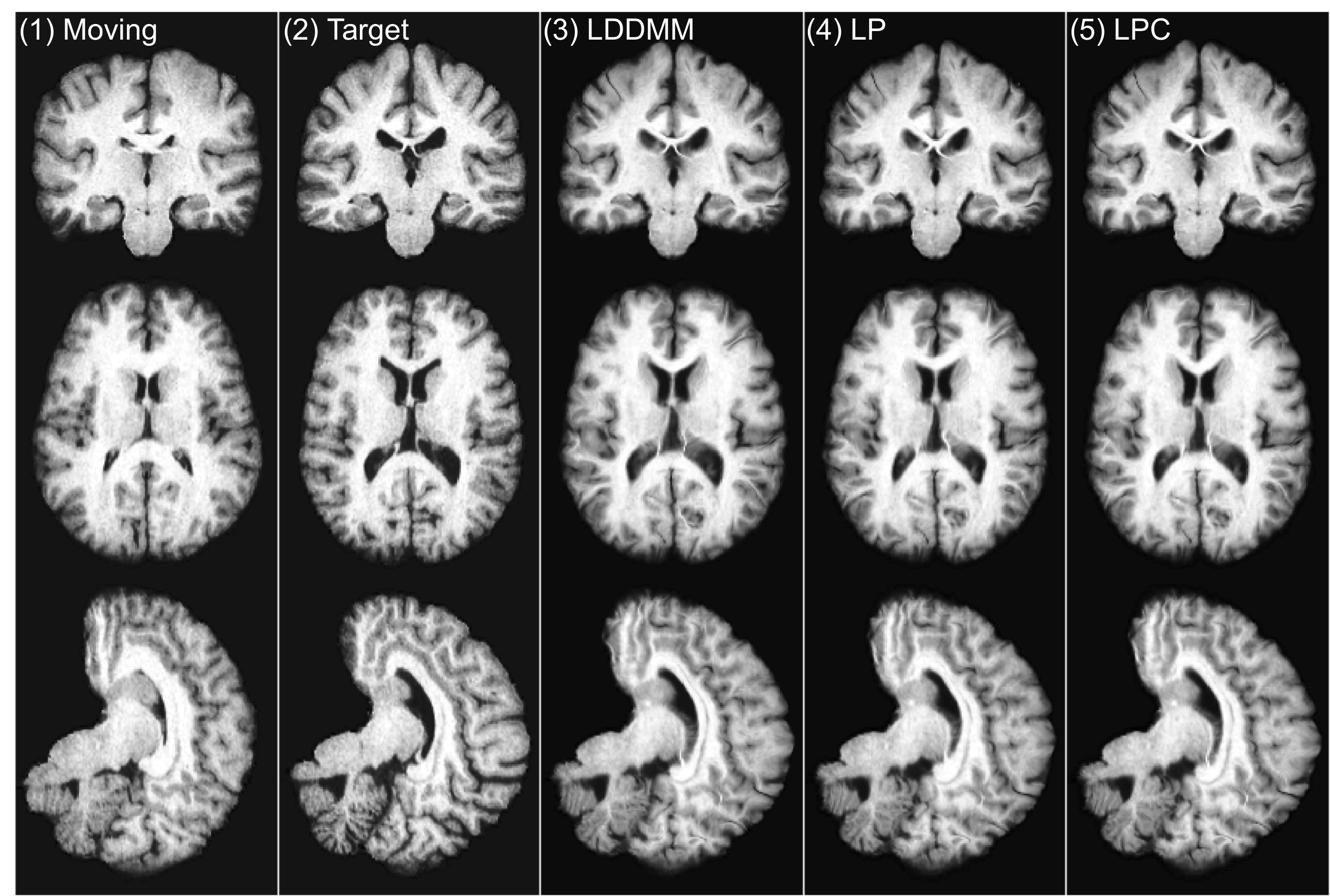}
        \caption{\texttt{LPBA40}}
    \end{subfigure}%
    ~ 
    \begin{subfigure}[t]{0.5\textwidth}
        \centering
        \includegraphics[width=3.5in]{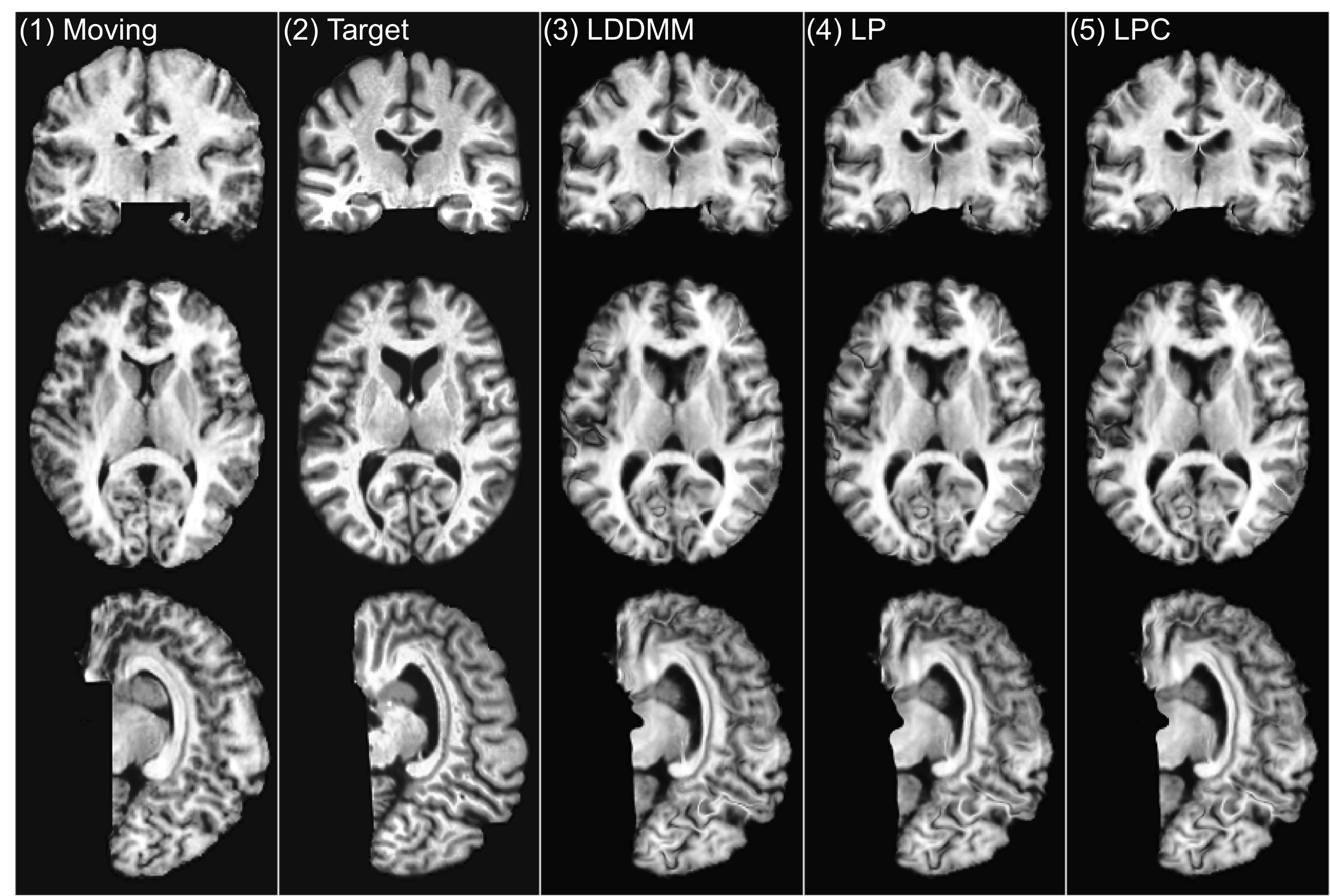}
        \caption{\texttt{IBSR18}}
    \end{subfigure}\\
	\vskip1.5ex
    \begin{subfigure}[t]{0.5\textwidth}
        \centering
        \includegraphics[width=3.5in]{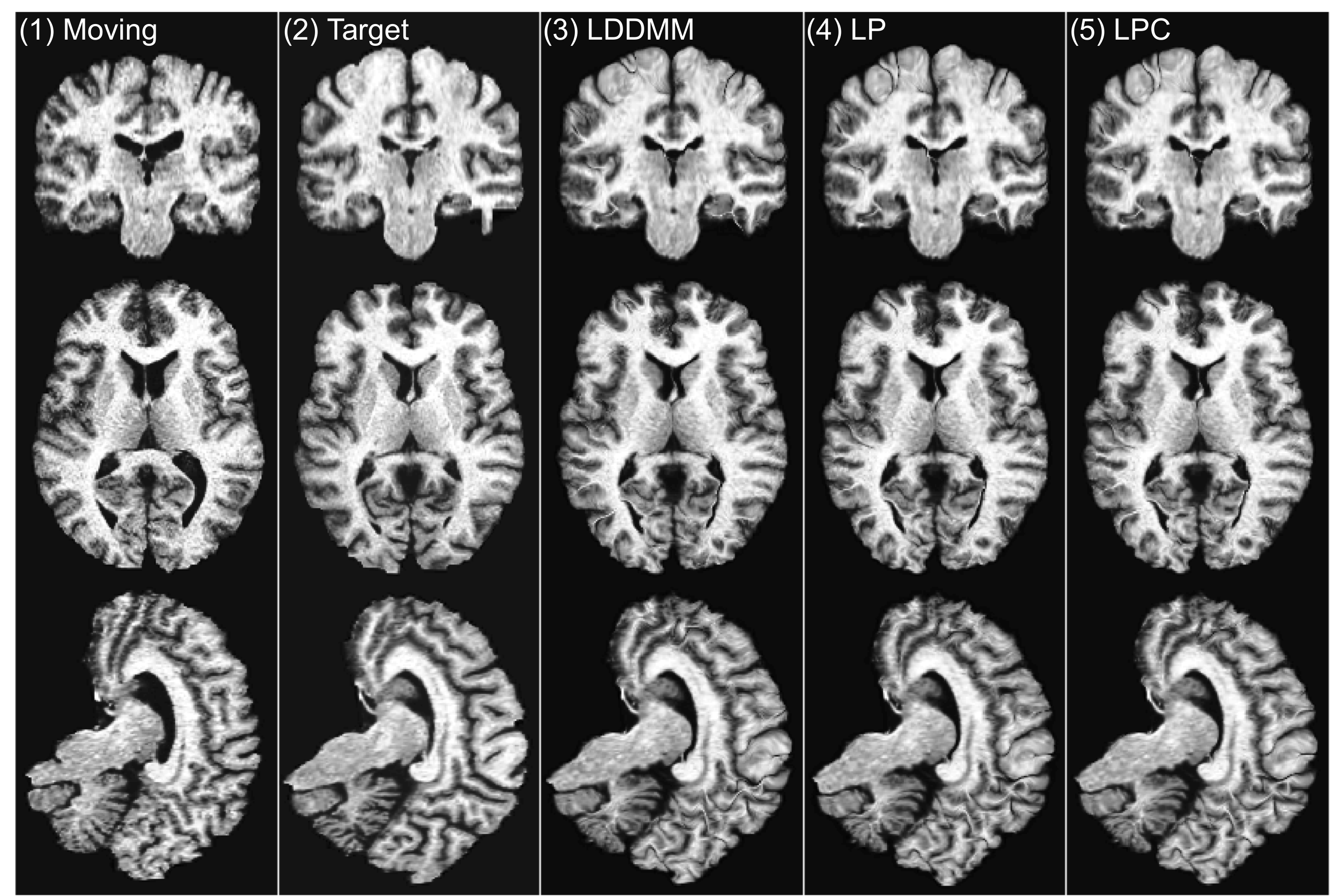}
        \caption{\texttt{CUMC12}}
    \end{subfigure}%
    ~ 
    \begin{subfigure}[t]{0.5\textwidth}
        \centering
        \includegraphics[width=3.5in]{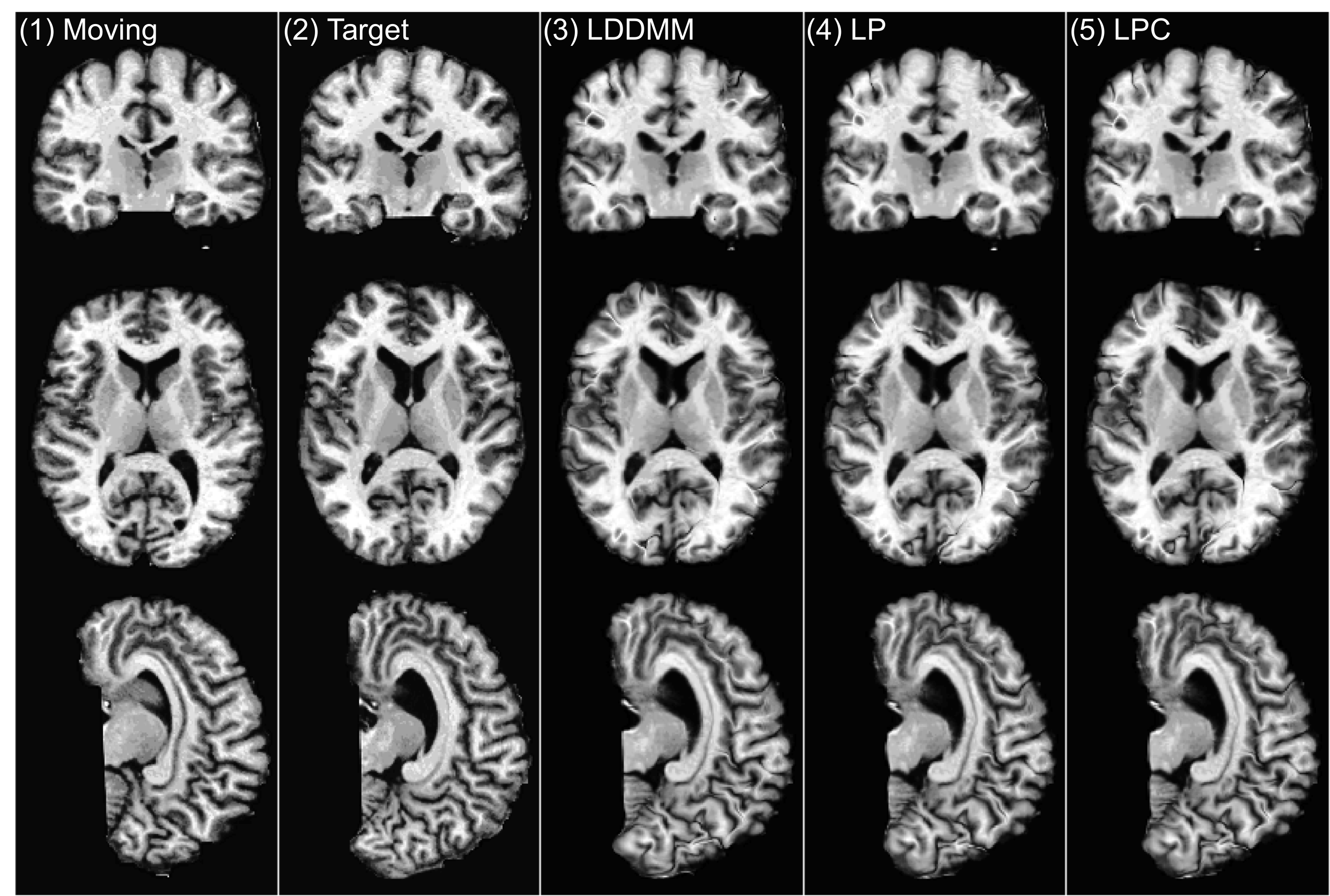}
        \caption{\texttt{MGH10}}
    \end{subfigure}
    \end{center}
    \caption{Example test cases for the \emph{image-to-image} registration. For every figure from \emph{left} to \emph{right}: (1): moving image; (2): target image; (3): registration result from optimizing LDDMM energy; (4): \xyrevision{registration result from prediction network (\texttt{LP})}; (5): registration result from prediction+correction network (\texttt{LPC}).}
    \label{fig:image_image_example}
\end{figure*}
\begin{table*}[h]
\footnotesize
\centering
\begin{tabularx}{\textwidth}{|r *{7}{|Y}|}
\hline
& \multicolumn{7}{c|}{\textbf{Deformation Error w.r.t LDDMM optimization on T1w-T1w data} [mm]}\\ \hline
\textsl{\xyrevision{Data percentile for all voxels}} & 0.3\% & 5\% & 25\% & 50\% & 75\% & 95\% & 99.7\%\\ \hline
Affine (Baseline) & 0.1664 & 0.46 & 0.9376 & 1.4329 & 2.0952 & 3.5037 & 6.2576\\ \hline
T1w-T1w \texttt{LP} & 0.0348 & 0.0933 &   0.1824 &   0.2726 &   0.3968 &   0.6779 &   1.3614 \\\hline
T1w-T1w \texttt{LPC} & 0.0289 &   0.0777 &   0.1536 &   0.2318 &   0.3398 &   0.5803 &   1.1584 \\\hline
T1w-T2w \texttt{LP} &  0.0544 &   0.1457 &   0.2847 &   0.4226 &   0.6057 &   1.0111 &   2.0402 \\\hline
T1w-T2w \texttt{LPC} & 0.0520 &   0.1396 &   0.2735 &   0.4074 &   0.5855 &   0.9701 &   1.9322 \\\hline
T1w-T2w \texttt{LP}, 10 images & 0.0660 &   0.1780 &   0.3511 &   0.5259 &   0.7598 &   1.2522 &   2.3496\\\hline
T1w-T2w \texttt{LPC}, 10 images & 0.0634 &   0.1707 &   0.3356 &   0.5021 &   0.7257 &   1.1999 &   2.2697\\\hline
\end{tabularx}
\vskip4ex
\caption{\xyrevision{Evaluation result for \textsl{multi-modal image-to-image} tests. \xyrevisionsecond{Deformation error (2-norm) per voxel between predicted deformation and optimization deformation. Percentiles over all deformation errors are shown to illustrate the error distribution.} \texttt{LP}: prediction network. \texttt{LPC}: prediction+correction network. 10 images: network is trained using 10 images (90 registrations as training cases).}}
\label{table:multimodal}
\end{table*}
\xyrevision{
\subsection{Multi-modal image registration}
In this task, a sliding window stride of 14 is used for the test cases. Table~\ref{table:multimodal} shows the prediction results compared to the deformation results obtained by T1w-T1w LDDMM optimization. The multi-modal networks (T1w-T2w, \texttt{LP}/\texttt{LPC}) significantly reduce deformation errors compared to affine registration, and only suffer a slight loss in accuracy compared to their T1w-T1w counterparts. This demonstrates the capability of our network architecture to implicitly learn the complex similarity measure between two modalities. Furthermore, for the networks trained using only 10 images, the performance only decreases slightly in comparison with the T1w-T2w multi-modal networks trained with 359 images. Hence, even when using very limited image data, we can still successfully train our prediction networks when a sufficient number of patches is available. Again, using a correction network improves the prediction accuracy in all cases. Fig.~\ref{Fig:multimodal} shows one multi-modal registration example. All three networks (T1w-T1w, T1w-T2w, T1w-T2w using 10 training images) generate warped images that are similar to the LDDMM optimization result.
\begin{figure*}[htbp]
\centering
\includegraphics[width=\textwidth]{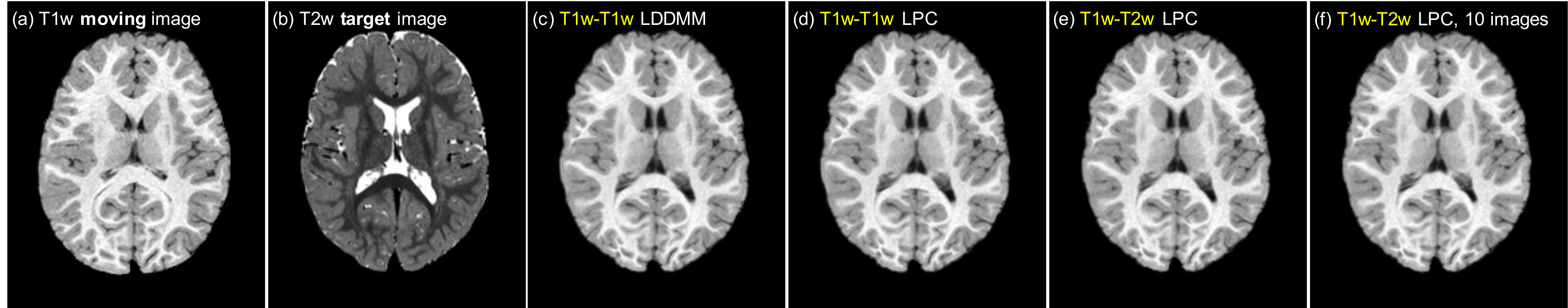}
\vskip4ex
\caption{\xyrevision{Example test case for \textsl{multi-modal image-to-image} tests. (a): T1w moving image; (b): T2w target image; (c): T1w-T1w LDDMM optimization (\texttt{LO}) result; (d)-(f): deformation prediction+correction (\texttt{LPC}) result using (d) T1w-T1w data; (e) T1w-T2w data; (f) T1w-T2w data using only 10 images as training data.}
\label{Fig:multimodal}}
\end{figure*}

}

\subsection{Runtime study}
\label{sec:runtime}
We assess the runtime of \texttt{Quicksilver} on a single Nvidia TitanX (Pascal) GPU. 
Performing LDDMM optimization using the GPU-based implementation of \texttt{PyCA} for a $229\times 193 \times 193$ 3D brain image takes approximately 10.8 minutes. Using our prediction network \xyadd{with a sliding window stride of 14}, the initial momentum prediction time is, on average, 7.63 seconds. Subsequent geodesic shooting to generate the deformation field takes 8.9 seconds, resulting in a total runtime of 18.43 seconds. Compared to the LDDMM optimization approach, our method achieves a $35\times$ speed up. Using the correction network together with the prediction network doubles the computation time, but the overall runtime is still an order of magnitude faster than direct LDDMM optimization. Note that, at a stride of 1, computational cost increases about 3000-fold in 3D, resulting in runtimes of about $5 \nicefrac{1}{2}$ hours for 3D image registration (eleven hours when the correction network is also used). Hence the initial momentum parameterization, which can tolerate large sliding window strides, is essential for fast deformation prediction with high accuracy while guaranteeing diffeomorphic deformations. 

Since we predict the whole image initial momentum in a patch-wise manner, it is natural to extend our approach to a multi-GPU implementation by distributing patches across multiple GPUs. We assess the runtime of this parallelization strategy on a cluster with multiple Nvidia GTX 1080 GPUs; the initial momentum prediction result is shown in Fig.~\ref{fig:multiGPU}. As we can see, by increasing the number of GPUs, the initial momentum prediction time decreases from 11.23 seconds (using 1 GPU) to 2.41 seconds using 7 GPUs. However, as the number of GPUs increases, the communication overhead between GPUs becomes larger which explains why computation time does not equal to $11.23 / {\text{number of GPUs}}$ seconds. \xy{Also, when we increase the number of GPUs to 8, the prediction time slightly increases to 2.48s. This can be attributed to the fact that \texttt{PyTorch} is still in Beta-stage and, according to the documentation, better performance for large numbers of GPUs (8+) is being actively developed\footnote{\url{http://pytorch.org/docs/master/notes/cuda.html\#use-nn-dataparallel-instead-of-multiprocessing}}. Hence, we expect faster prediction times using a large number of GPUs in the future}. Impressively, by using multiple GPUs, the runtime can be improved by two orders of magnitude over a direct (GPU-based) LDDMM optimization. Thus, our method can readily be used in a GPU-cluster environment for ultra-fast deformation prediction.
\begin{figure}[t!]
\begin{center}
\includegraphics[width=0.4\textwidth]{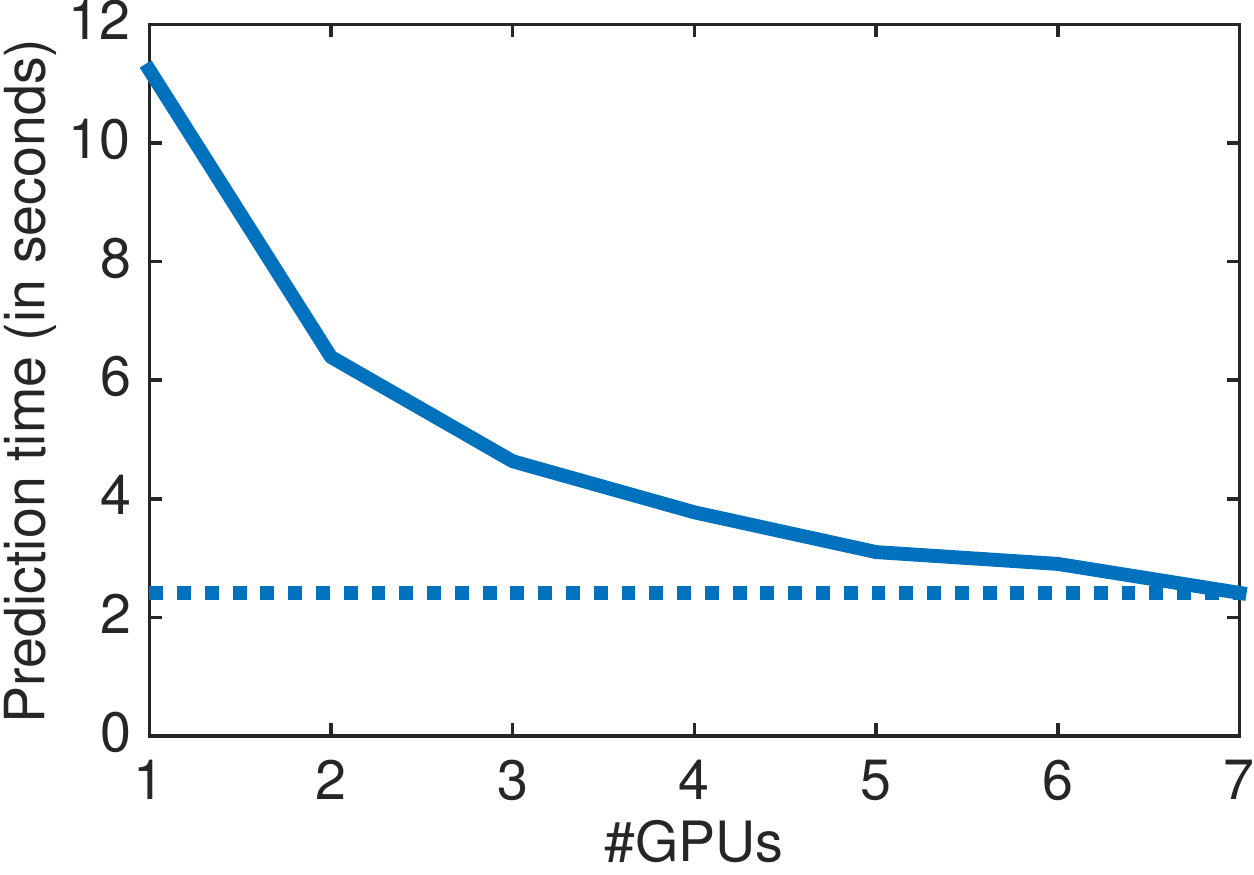}
\end{center}
\caption{Average initial momentum prediction time (in seconds) for a single $229\times 193 \times 193$ 3D brain image case using various number of GPUs.}
\label{fig:multiGPU}
\end{figure}
\section{Discussion}
\label{sec:discussion}
We proposed a fast registration approach based on the patch-wise prediction of the initial momentum parameterization of the LDDMM shooting formulation. The proposed approach allows taking large strides for patch-wise prediction, without a substantial decrease in registration accuracy, resulting in fast and accurate deformation prediction. The proposed correction network is a step towards highly accurate deformation prediction, while only decreasing the computation speed by a factor of 2. Our method retains all theoretical properties of LDDMM and results in diffeomorphic transformations if appropriately regularized, but computes these transformations an order of magnitude faster than a GPU-based optimization for the LDDMM model. Moreover, the patch-wise prediction approach of our methods enables a multi-GPU implementation, further increasing the prediction speed. In effect, our \texttt{Quicksilver} registration approach converts a notoriously slow and memory-hungry registration \rknew{approach} to a fast method, while retaining all of its appealing mathematical properties.

\vskip1ex
Our framework is very general and can be directly applied to many other registration techniques. For non-parametric registration methods with pixel/voxel wise registration parameters (e.g., elastic registration~\cite{modersitzki2004}, or stationary velocity field~\cite{vercauteren2009} registration approaches), our approach can be directly applied for parameter prediction. For parametric registration methods with local control such as B-splines, we could attach fully connected layers to the decoder to reduce the network output dimension, thereby predicting low-dimensional registration parameters for a patch. Of course, the patch pruning techniques may not be applicable for these methods if the parameter locality cannot be guaranteed.

\vskip1ex
In summary, the presented deformation prediction approach is the first step towards more complex tasks where fast, deformable, predictive image registration techniques are required. It opens up possibilities for various extensions and applications. Exciting possibilities are, for example, to use \texttt{Quicksilver} as the registration approach for fast multi-atlas segmentation, fast image geodesic regression, fast atlas construction, or fast user-interactive registration refinements (where only a few patches need to be updated based on local changes). Furthermore, extending the deformation prediction network to more complex registration tasks \rknew{could also be} beneficial;  \mnr{e.g., to further explore the behavior of the prediction models for multi-modal image registration~\cite{yang2017multimodal}.} Other potential areas include joint image-label registration for better label-matching accuracy; multi-scale-patch networks for very large deformation prediction; deformation prediction for registration models with anisotropic regularizations; and end-to-end optical flow prediction via initial momentum parameterization. Other correction methods could also be explored, by using different network structures, or by recursively updating the deformation parameter prediction using the correction approach \xyrevisionsecond{(e.g., with a sequence of correction networks where each network corrects the momenta predicted from the previous one)}. Finally, since our uncertainty quantification approach indicates high uncertainty for areas with large deformation or appearance changes, utilizing the uncertainty map to detect pathological areas could also be an interesting research direction.

\vskip1ex 
\noindent
\textbf{Source code.}
\xyadd{To make the approach readily available to the community, we open-sourced \texttt{Quicksilver} at \url{https://github.com/rkwitt/quicksilver}.} \mn{Our long-term goal is to make our framework the basis for different variants of predictive image registration; e.g., to provide \texttt{Quicksilver} variants for various organs and imaging types, as well as for different types of spatial regularization.}

\vskip1ex 
\noindent
\textbf{Acknowledgments.}
This work is supported by NIH 1 R41 NS091792-01, \xyrevision{NIH HDO55741}, NSF
ECCS-1148870, and EECS-1711776. \xyrevision{This work also made use of the XStream computational resource, supported by the National Science Foundation Major Research Instrumentation program (ACI-1429830).} We also thank Nvidia for the donation of a TitanX GPU.

\section*{References}
\bibliography{mybibfile}

\end{document}